\documentclass[journal]{IEEEtran}

\usepackage{xcolor,soul,framed} 

\colorlet{shadecolor}{yellow}
\usepackage[pdftex]{graphicx}
\graphicspath{{../pdf/}{../jpeg/}}
\DeclareGraphicsExtensions{.pdf,.jpeg,.png}

\usepackage[cmex10]{amsmath}
\usepackage{array}
\usepackage{mdwmath}
\usepackage{mdwtab}
\usepackage{eqparbox}
\usepackage{url}
\usepackage{lipsum}
\usepackage[noadjust]{cite}
\usepackage{caption}
\usepackage{cleveref}
\usepackage{subcaption}
\usepackage{amsmath}
\usepackage{amssymb}
\usepackage{multirow}

\usepackage{graphicx}
\usepackage{xcolor}

\usepackage{microtype}
\usepackage{booktabs}

\DeclareMathOperator{\E}{\mathbb{E}}

\begin{document}
\title{PoE: a Panel of Experts for Generalized Automatic Dialogue Assessment}
  \author{Chen Zhang,~\IEEEmembership{Student Member,~IEEE,}
      Luis Fernando D'Haro,~\IEEEmembership{Member,~IEEE,} \\
      Qiquan Zhang,~\IEEEmembership{Member,~IEEE,}
      Thomas Friedrichs,~\IEEEmembership{Member,~IEEE,}
      Haizhou Li,~\IEEEmembership{Fellow,~IEEE}

  \thanks{Chen Zhang, Qiquan Zhang, and Haizhou Li are with the Human Language Technology Group at Electrical \& Computer Engineering Department of National University of Singapore, Singapore (email: chen\_zhang, haizhou.li@u.nus.edu, qiquanzh@nus.edu.sg).}%
  \thanks{Luis Fernando D’Haro is with the Speech Technology Group of Universidad Politécnica de Madrid (ETSIT, UPM), Spain (email: luisfernando.dharo@upm.es).}
  \thanks{Thomas Friedrichs is with Robert Bosch (SEA) Pte Ltd, Singapore (email: Thomas.Friedrichs@sg.bosch.com).}
  \thanks{Haizhou Li is with the Guangdong Provincial Key Laboratory of Big Data Computing, The Chinese University of Hong Kong, Shenzhen, China (email: haizhouli@cuhk.edu.cn)}
  
  }  

\markboth{IEEE/ACM Transactions on Audio Speech and Language Processing, manuscript, MARCH~2022 (Under review)
}{xxx \MakeLowercase{\textit{et al.}}: PoE: a Panel of Experts for Generalized Automatic Dialogue Assessment}

\maketitle

\begin{abstract}
Chatbots are expected to be knowledgeable across multiple domains, e.g. for daily chit-chat, exchange of information, and grounding in emotional situations. 
To effectively measure the quality of such conversational agents, a model-based automatic dialogue evaluation metric (ADEM) is expected to perform well across multiple domains. Despite significant progress, an ADEM that works well in one domain does not necessarily generalize to another. This calls for a dedicated network architecture for domain generalization. To tackle the multi-domain dialogue evaluation task, we propose a \textit{Panel of Experts} (PoE), 
a multitask network that consists of a shared transformer encoder and a collection of lightweight adapters. The shared encoder captures the general knowledge of dialogues across domains, while each adapter specializes in one specific domain and serves as a domain expert. To validate the idea, we construct a high-quality multi-domain dialogue dataset leveraging data augmentation and pseudo labeling. The PoE network is comprehensively assessed on 16 dialogue evaluation datasets spanning a wide range of dialogue domains. It achieves state-of-the-art performance in terms of mean Spearman correlation over all the evaluation datasets. It exhibits better zero-shot generalization than existing state-of-the-art ADEMs and the ability to easily adapt to new domains with few-shot transfer learning.

\end{abstract}


%
\IEEEpeerreviewmaketitle


\section{Introduction}
\label{sec:introduction}
\IEEEPARstart{T}{he} research advancement on open-domain dialogue systems, a.k.a. chatbots is guided by evaluation. The evaluation of chatbots is a complex task as the conversations carried out by chatbots can be about any topic, and of very different characteristics, such as daily chit-chat~\cite{li-etal-2017-dailydialog}, knowledge exchange~\cite{gopalakrishnan2019topical}, emotion disclosure~\cite{rashkin-etal-2019-towards}, and personal interests~\cite{zhang-etal-2018-personalizing}. Especially, as chatbots are increasingly expected to perform in multiple domains~\cite{shuster-etal-2020-dialogue,roller-etal-2021-recipes}, the corresponding evaluation methods ought to be equally versatile. While human judges have no issue in assessing such a wide range of topics given proper instructions, it is too costly to perform human evaluation at every stage of system development~\cite{mehri-eskenazi-2020-usr}. This prompts us to develop an automatic evaluation metric, which highly correlates with human evaluation under different evaluation scenarios. 

Recently, there is a rising interest in model-based reference-free automatic dialogue evaluation metrics (ADEMs), that has advantage over the commonly used reference-based untrained metrics such as BLEU~\cite{papineni-etal-2002-bleu} and F-score, which are shown to correlate poorly w.r.t. human evaluations~\cite{liu-etal-2016-evaluate}. Most of the reference-free ADEMs are trained on human-human dialogue corpora in a weakly supervised fashion. Specifically, a model is trained to classify a dialogue response as either positive or negative given its dialogue context\footnote{Sentences irrelevant, semantically inappropriate or incoherent w.r.t. a dialogue context can serve as negative responses.}, which consists of several consecutive utterances from a human-human dialogue. During training, a true dialogue response in the context is considered as a positive response, whereas the negative responses are obtained via different semantic or syntactic perturbation strategies~\cite{sinha-etal-2020-learning,zhang-etal-2021-dscore}. During inference,  the trained model is used to score responses given their dialogue contexts.

The recent model-based ADEMs~\cite{ghazarian-etal-2019-better,lan-etal-2020-pone,mehri-eskenazi-2020-usr,huang-etal-2020-grade,sinha-etal-2020-learning,zhang-etal-2021-dscore} have demonstrated a strong correlation with human evaluation on different dialogue evaluation datasets. However, their generalizability across different dialogue domains is questionable. A recent survey~\cite{yeh-etal-2021-comprehensive} shows that the state-of-the-art ADEMs  obtain fair in-domain performance, i.e., good correlations on dialogue data similar to their training data. However, when evaluated on evaluation data different from their training data, the ADEMs tend to perform poorly. This issue has been frequently raised in recent works on dialogue evaluation~\cite{zhang2022mddeval,zhao2022floweval,smith2022human}. In addition, most ADEMs only consist of a single network, for example, a pre-trained transformer encoder, such as BERT~\cite{devlin-etal-2019-bert} or RoBERTa~\cite{liu2019roberta} with a classification layer on top. They don't employ a specific mechanism for domain generalization. Hence, it is believed that an adequate network architecture is required for multi-domain dialogue evaluation.

\begin{figure*}[ht!]
\flushright
\includegraphics[width=0.9\linewidth]{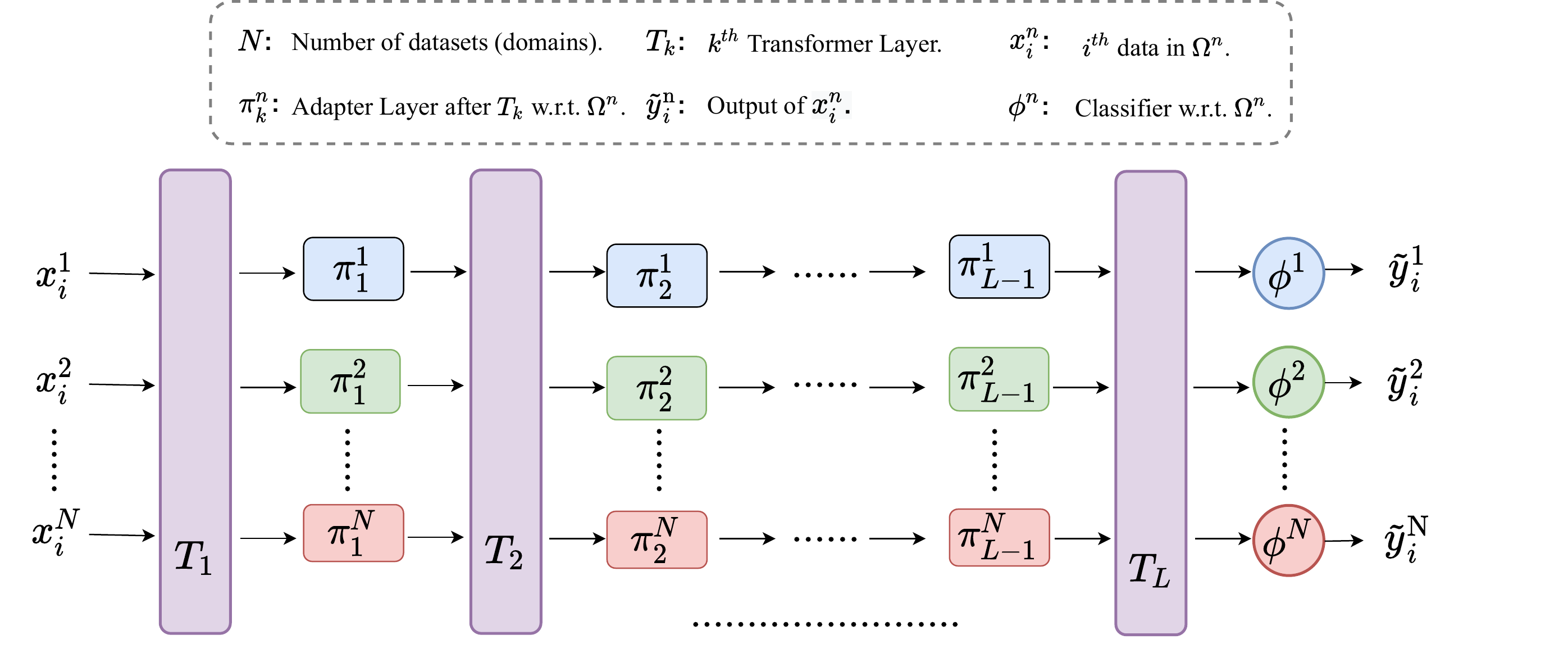} 
\caption{System architecture of a Panel of Experts (PoE). A transformer encoder $T$ consists of $L$ layers (purple rectangles). Different colors (blue, red, and green) denote domain-specific modules. $\{\pi^n\}_{n=1}^{|N|}$ are the N different domain-specific adapters. Each domain-specific $\pi^n$ has $L-1$ adapter layers, $\{\pi^n_1, \pi^n_2,\ldots,\pi^n_{L-1}\}$, that are injected in between every two consecutive transformer layers. $\{\phi^n\}_{n=1}^{|N|}$ are the domain-specific classifiers after the final transformer layer, $T_L$. $T$ is shared by all the domain-specific modules.} 
\label{fig:system}
\end{figure*}

We propose a network architecture as a single-model metric with a mechanism to handle domain generalization. The network makes a unified decision with multiple domain specific experts, thus is referred to as a Panel of Experts (PoE). It consists of a pretrained transformer encoder~\cite{devlin-etal-2019-bert,liu2019roberta} and a set of adapters~\cite{pmlr-v97-houlsby19a} (Figure~\ref{fig:system}) which is shared across domains. The adapters are lightweight
task-specific modules interleaved between the layers of the pretrained transformer encoder. Each adapter serves as a domain expert in evaluating a specific category of dialogues. PoE is also flexible when performing out-of-domain evaluation tasks. For instance, we can either average the prediction scores of all adapters (late fusion) or average the parameters of the adapters to derive a single adapter (early fusion) for decision making. Furthermore, we may adapt PoE to new domain with few-shot transfer learning, without the need of full model training or finetuning.



To provide a high-quality multi-domain dataset for this study, we construct a training dataset from five commonly-used and high-quality human-human dialogue corpora leveraging data augmentation and pseudo labeling~\cite{lee2013pseudo}. We compare our PoE metric with the state-of-the-art metrics and two strong baselines trained on our constructed multi-domain dataset. 

In this paper, we make the following contributions: (1) We bridge the gap between existing model-based reference-free ADEMs and a strong multi-domain automatic dialogue evaluation metric, which can also effectively handle out-of-domain evaluation. More specifically, we realize this by constructing a high-quality multi-domain dataset for training the ADEMs and proposing PoE, a novel automatic dialogue evaluation metric based on transformer adapters. To our knowledge, PoE is the first multitask model that targets evaluation across a wide range of dialogue domains. (2) We empirically show that PoE outperforms existing state-of-the-art ADEMs as well as strong baselines on a large collection of evaluation datasets that covers different dialogue domains. In total, there are 11 in-domain and 5 out-of-domain evaluation datasets. (3) The implementation of PoE, datasets, and pretrained checkpoints will be released to the public, allowing researchers and practitioners to use or adapt them for their own evaluation tasks.

The remaining sections are organized as follows: \S\ref{sec:related_work} discusses the related work. In \S\ref{sec:problem}, we formally define the dialogue evaluation task. In \S\ref{sec:mese}, we explain the proposed PoE metric. \S\ref{sec:dataset} describes the methods to construct the multi-domain dialogue dataset for training automatic dialogue evaluation metrics. The experiment preliminaries are outlined in \S\ref{sec:exp-setup}. \S\ref{sec:exp-results} presents the experimental results and detailed analyses, which include in-domain, out-of-domain, and few-shot transfer analysis. The last section concludes the paper and outlines the future work.  

\section{Related Work}
\label{sec:related_work}

\subsection{Automatic Dialogue Evaluation Metrics (ADEMs)}
There are a number of commonly used evaluation metrics, that are simple, reference-based, and non-trainable. One category is the word-overlap metrics, such as BLEU~\cite{papineni-etal-2002-bleu}, ROUGE~\cite{lin-2004-rouge}, and METEOR~\cite{banerjee-lavie-2005-meteor}. This category of metrics assigns a score to the dialogue response based on its word or n-gram overlap with the corresponding human-written references. The other category is the embedding-based metrics, such as Greedy Matching~\cite{rus2012optimal} and Embedding Average~\cite{mitchell2008vector}. By leveraging static word vectors, the embedding-based metrics move beyond surface-level matches and focus more on the semantic similarities between a dialogue response and the corresponding references. Despite their simplicity, both categories are often criticized for their poor correlation with human evaluation~\cite{liu-etal-2016-evaluate}. The crux is that there are many possible responses to one given context in an open-ended dialogues~\cite{zhao-etal-2017-learning}.

Addressing the problem of multiple possible responses, the study of ADEMs shifts from reference-based approaches towards model-based reference-free ones~\cite{yeh-etal-2021-comprehensive}. Examples of recent model-based reference-free metrics include BERT-RUBER~\cite{ghazarian-etal-2019-better}, PONE~\cite{lan-etal-2020-pone}, USR-DR~\cite{mehri-eskenazi-2020-usr}, GRADE~\cite{huang-etal-2020-grade}, and MaUdE~\cite{sinha-etal-2020-learning}. There are several common characteristics among these ADEMs: (1) To evaluate whether or not a response is appropriate with respect to a dialogue context as opposed to one or more references. (2) To employ a pre-trained language model~\cite{devlin-etal-2019-bert,liu2019roberta} to improve the ADEMs' classification capability (3) To avoid human annotations by using context-response data derived from a single human-human dialogue corpus. Examples are DailyDialog~\cite{li-etal-2017-dailydialog}, PersonaChat~\cite{zhang-etal-2018-personalizing} or TopicalChat~\cite{gopalakrishnan2019topical}. Despite much success, the model-based reference-free approaches face several challenges. A major one is their inability to generalize to dialogue data beyond what they are trained on. 


In this paper, we propose to tackle the problem from both the algorithm and data perspectives by developing a novel ADEM based on transformer adapters~\cite{pmlr-v97-houlsby19a}.  We also construct a high-quality dataset to facilitate the study. Zhang et al.~\cite{zhang2022mddeval} proposed MDD-Eval which also targets multi-domain dialogue evaluation. However, there are several key differences between both works. First, the underlying algorithms are completely different. The MDD-Eval metric is a single model trained on pseudo-labeled augmented data. It lacks the mechanism to handle domain-specific datasets in a specialized manner. On the other hand, our PoE metric adopts multi-task learning with the parameter-efficient adapter network. Each adapter module serves as a domain expert. Second, even though the training data of both MDD-Eval and PoE are constructed in a similar manner, PoE is more compatible with the data collection pipeline, because it handles increasing number of datasets more efficiently. MDD-Eval or single-model metrics in general require retraining or full-model finetuning when adapting to new domain-specific or task-specific data, whereas for PoE, we can just add new parameter-efficient adapter modules to handle the new datasets. \S\ref{subsec:apply-drs} presents the empirical evidence to support our claim. Third, in \S\ref{sec:exp-results}, we show that PoE is a much stronger multi-domain evaluation metric than MDD-Eval under the in-domain, out-of-domain and few-shot transfer settings.


It is noted that dialogue quality is multi-faceted in nature~\cite{mehri-eskenazi-2020-usr}, model-based ADEMs are often designed for response appropriateness, as well as engagement~\cite{Ghazarian_Weischedel_Galstyan_Peng_2020}, naturalness~\cite{mehri-eskenazi-2020-usr}, adequacy~\cite{zhang2020deep}, coherence~\cite{dziri-etal-2019-evaluating,zhang-etal-2021-dynaeval}, consistency~\cite{welleck-etal-2019-dialogue,nie-etal-2021-like}, etc. There have been studies on combining different specialized models for multi-dimensional\footnote{Multi-dimension refers to multiple dialogue qualities.} dialogue evaluation, for example, D-score~\cite{zhang-etal-2021-dscore}, HolisticEval~\cite{pang-etal-2020-towards}, USR~\cite{mehri-eskenazi-2020-usr}, and USL-H~\cite{phy-etal-2020-deconstruct}. As the scope of this paper is on multi-domain dialogue evaluation rather than multi-dimensional dialogue evaluation, we focus on dialogue qualities that are more frequently studied in the literature and highly correlate with the pre-training objective of our proposed metric, the response appropriateness.


\subsection{Multitask Learning}

In multitask learning~\cite{crawshaw2020multi}, a model is trained simultaneously with multiple tasks and a shared representation is learned to capture the commonalities among the related tasks~\cite{crawshaw2020multi}. Multitask learning is an effective approach to reduce over-fitting to a particular task and thus, improve generalizability of the models~\cite{caruana1997multitask,ruder2017overview}. It has been successfully applied in a wide range of natural language processing tasks, such as semantic parsing~\cite{9172116}, sequence labeling~\cite{9153102}, language modeling~\cite{devlin-etal-2019-bert,radford2019language}, machine translation~\cite{luong2015multi,wang-etal-2020-multi}, and dialogue~\cite{rastogi-etal-2018-multi,9376902}. 

In the context of open-domain dialogue evaluation, we recently applied  multitask learning~\cite{zhang-etal-2021-dscore} for a holistic assessment of dialogues whereby the related tasks are designed to evaluate different dialogue qualities, including language fluency, coherence, semantic appropriateness, and logical consistency. Unlike~\cite{zhang-etal-2021-dscore}, which addresses multi-dimensional evaluation, we study multi-domain evaluation in this paper by designing a hard-parameter sharing network, which consists of multiple transformer adapters sharing an underlying pretrained transformer encoder. Through multitask learning, the shared transformer encoder is expected to adapt and capture the general knowledge of dialogues while the domain-specific adapters capture specific properties with respect to the respective domains. 

\subsection{Adapters}

Adapters are lightweight task-specific modules interleaved between layers of a pre-trained network~\cite{NIPS2017_e7b24b11,pmlr-v97-houlsby19a}. Adapter-based transfer learning performs similarly to full finetuning, but being more parameter efficient. Houlsby et al.~\cite{pmlr-v97-houlsby19a} demonstrates that on the GLUE benchmark~\cite{wang2018glue}, finetuning only the task-specific adapters attains within 0.4\% of the performance of full fine-tuning, which requires adapting all parameters of a pretrained model for each task. In a separate study, Stickland and Murray~\cite{pmlr-v97-stickland19a} propose a neural architecture that adds task-specific adapters to BERT~\cite{devlin-etal-2019-bert} and train the entire network with a multitask learning setup. This neural architecture performs similarly to those separately finetuned on the GLUE benchmark. 


More recently, Friedman et al.~\cite{friedman-etal-2021-single} proposes the MADE model for extractive question and answering tasks. MADE consists of a shared transformer encoder, dataset-specific token classifiers and adapters. When training on a mixture of source datasets, all parameters within MADE are jointly optimized. The model has attained strong in-domain and out-of-domain performance. Following this line of thought, we apply the adapter-based multitask network in the multi-domain dialogue evaluation for the first time. 

\subsection{\textcolor{black}{Ensemble}}

\textcolor{black}{Ensemble is a common technique for boosting prediction accuracy. The conventional ensemble involves two steps: (1) making predictions with multiple independent models; (2) integrating the predictions into a final result~\cite{dong2020survey}. In open-domain dialogue evaluation, prior works, such as USR~\cite{mehri-eskenazi-2020-usr}, USL-H~\cite{phy-etal-2020-deconstruct}, and D-score~\cite{zhang-etal-2021-dscore}, ensemble multiple metrics to boost correlation with the overall human judgment. For PoE, we apply the unweighted average of predictions inferred by different domain-specific adapters to obtain the final prediction score and examine whether such an approach can yield good out-of-domain performance.}

\textcolor{black}{Besides the conventional ensemble of model predictions, Friedman et al.~\cite{friedman-etal-2021-single} propose a simple method to average the parameters of multiple adapters, which achieves good generalization on unseen question answering datasets. Matena \& Raffel~\cite{Matena2021merging} propose to merge pre-trained language models that are fine-tuned on various text classification tasks via parameter averaging. In the same line of thought, we assess whether parameter averaging of the domain-specific adapters can achieve performance similar to that of PoE while incurring less inference and memory cost. Concurrent to our work, Wortsman et al.~\cite{pmlr-v162-wortsman22a} explore averaging the weights of multiple models that are fine-tuned with different hyper-parameter configurations on the same task. They demonstrate that simple parameter averaging attains strong performance in image classification. Their work further validates the effectiveness of parameter averaging.}


\section{Problem Formulation}
\label{sec:problem}

In this section, we formally define the multi-domain dialogue evaluation task. Assume we have a collection of $J$ dialogue evaluation datasets, denoted as $\mathcal{D}$. An evaluation dataset 
within $\mathcal{D}$ is denoted as $D^j$, where $j\in\{1,...,J\}$.

$D^j$ contains $I$ number of dialogue context-response pairs\footnote{The context is one or a few consecutive utterances drawn from a human-human dialogue, and the response is generated by a chatbot conditioning on the context.}. We denote the context and the corresponding response as $c_{i}^j$ and $r_{i}^j$ respectively, where $i \in\{1,...,I\}$. In addition, each $(c_{i}^j, r_{i}^j)$ is annotated by several human judges, and each human judge will provide a quality score based on the Likert scale to indicate his/her perception of the quality of $(c_{i}^j, r_{i}^j)$. The mean human score w.r.t. $(c_{i}^j, r_{i}^j)$ is denoted as $q_{i}^j$. 

We aim to learn a metric, $M(c_{i}^j, r_{i}^j) \rightarrow s_{i}^j$ where $s_{i}^j$ is the metric score that reflects the quality of $(c_{i}^j, r_{i}^j)$ as perceived by $M$. To assess the performance of $M$ on $D^j$, the correlation score between $S^j = \{s_{1}^j,\ldots,s_{I}^j\}$ and $Q^j = \{q_{1}^j,\ldots,q_{I}^j\}$ are computed. We use $\rho^j$ to represent the correlation score on $D^j$. Higher $\rho^j$ indicates better performance on $D^j$. To test the performance of $M$ on the $J$ evaluation datasets, we compute the average correlation $\tilde{\rho} = \frac{1}{J}\sum^{J}_{j=1}{\rho^j}$. \textcolor{black}{The specific form of correlation we adopt is the Spearman's rank correlation~\cite{daniel1987spearman}, a common statistical measure used for assessing metrics' performance. Spearman’s rank correlation determines the monotonic relationship between two variables. It is appropriate for continuous and discrete ordinal variables~\cite{corder2014nonparametric}. In our case,
both $S^j$ and $Q^j$ are treated as two continuous variables and their Spearman's rank correlation can be computed as follows:}

\begin{equation*}
\textcolor{black}{
    \rho^j = 1 - \frac{6\sum{\mathrm{k}^j_i}^2}{I(I^2-1)}
}
\end{equation*}
\textcolor{black}{where $k^j_i$ is the difference between the ordinal rank of $s_{i}^j$ within $S^j$ and that of $q_{i}^j$ within $Q^j$. The range of Spearman's rank correlation is between +1 and -1. +1 means a perfect monotonic association of the two variables. -1 means a perfect negative monotonic association of the two variables. 0 means that there is no association between the two variables.}

Suppose that we train $M$ with a collection of $N$ training datasets, denoted as $\Omega$. Each dataset in $\Omega$ belongs to a unique dialogue domain, $n$. Hence, $\Omega$ covers $N$ different domains. We denote each training dataset as $\Omega^n$ where $n\in\{1,...,N\}$. 


We want to assess both the in-domain and out-of-domain performance of $M$. For in-domain assessment, $\tilde{\rho}$ is computed over the subset of $\mathcal{D}$, of which the data are in-distribution w.r.t. $\Omega^n$. For out-of-domain assessment, $\tilde{\rho}$ is computed over the subset of $\mathcal{D}$, which contain out-of-distribution data w.r.t. $\Omega$.
 
\section{A Panel of Experts}
\label{sec:mese}

\subsection{\textcolor{black}{Motivation of A Panel of Experts}}
\label{subsec:sys-arc-motivation}

\textcolor{black}{Research on domain generalization remains under-explored in the field of automatic dialogue evaluation. According to a recent survey on domain generalization~\cite{9782500}, there are mainly three method categories to improve domain generalization. First, data manipulation, such as data augmentation and data generation. Second, representation learning, such as domain invariant representation learning and feature disentanglement. Third, learning strategies, such as ensemble learning and multi-task learning. To our knowledge, only the recent work, MDD-Eval~\cite{zhang2022mddeval} targets domain generalization in automatic dialogue evaluation. The key idea of MDD-Eval falls under the first method category. Yet, purely relying on multi-domain training data is not enough for domain generalization. In fact, the above-mentioned method categories are complementary to each other and can be combined towards better performance~\cite{9782500}. This motivates us to study better learning strategies and network architecture that can further enhance domain generalization of model-based metrics.} 

\textcolor{black}{In terms of learning strategy, multi-task training is a natural choice to improve domain generalization. Through a shared representation, the learning on a particular domain-specific dataset can be improved by leveraging additional information from other related domains~\cite{caruana1997multitask}. Furthermore, the model needs to adapt to new dialogues that are unseen during the training process. It is often infeasible to re-train the model or conduct full-model fine-tuning whenever there are new dialogue data. Therefore, we need light-weight dedicated network to adapt to the new dialogues and meanwhile, reuse the general knowledge gained in the pre-training process. We also want to reduce catastrophic forgetting that may happen during full-model fine-tuning. Hence, the adapter~\cite{pmlr-v97-houlsby19a} comes as a feasible choice.}

\textcolor{black}{Although multi-task learning and adapters have been proven effective in other tasks, such as text classification~\cite{pmlr-v97-stickland19a}, image classification~\cite{NIPS2017_e7b24b11}, and cross-lingual transfer~\cite{pfeiffer-etal-2020-mad}, to our knowledge, we are the first to apply them in automatic dialogue evaluation and comprehensively analyze their effectiveness in terms of in-domain, out-of-domain, and few-shot transfer evaluation across a diverse set of evaluation benchmarks.}

\subsection{System Architecture}
\label{subsec:sys-arc}

As shown in Figure~\ref{fig:system}, we formulate a Panel of Experts (PoE) as an automatic dialgoue evaluation metric, which consists of a pretrained transformer encoder, $N$ adapters, and $N$ classifiers. The transformer encoder, denoted as $T$, contains $L$ number of layers, $\{T_1,\ldots,T_L\}$. Each adapter, denoted as $\pi^n$, consists of a series of adapter layers, $\{\pi_1^n,\pi_2^n,\ldots,\pi_{L-1}^n\}$. An adapter layer is interleaved between two consecutive transformer layers. For example, $\pi_{L-1}^n$ is inserted in between $T_{L-1}$ and $T_L$. Each classifier, denoted as $\phi^n$, is a single-layer feed-forward network followed by a sigmoid activation function. As defined in \S\ref{sec:problem}, we have $N$ training datasets with each cover a unique dialogue domain. $\{\pi^n, \phi^n\}$ learn to classify responses from their respective domain-specific dataset $\Omega^n$. During training, all $\{\pi^n, \phi^n\}_{n=1}^{|N|}$ and $T$ are jointly optimized in a multitask learning manner.

The input data of PoE, $x_i^n$ is a context-response pair from $\Omega^n$. The pair is denoted as $(c_i^n, r_i^n)$. The associated label of $x_i^n$ is denoted as $y_i^n$, which indicates whether $r_i^n$ is appropriate w.r.t. $c_i^n$. It is either 1 (appropriate) or 0 (inappropriate). Note that $y_i^n$ is a pseudo label (discussed in \S\ref{subsec:pl}) instead of the ground-truth human label. Hence, PoE can be seen as a semi-supervised approach, which differs from all existing model-based dialogue evaluation metrics, except our MDD-Eval metric~\cite{zhang2022mddeval}. However, the architecture of MDD-Eval is simple (a pretrained RoBERTa~\cite{liu2019roberta} plus a feed-forward classification network) while PoE benefits from the transformer adapter and multitask learning for improved prediction ability and generalizability.

$(c_i^n, r_i^n)$ is concatenated into a single sequence of tokens including special start, end, and separation tokens when fed into the network: ``$<$s$>c_i^n<$/s$>r_i^n<$/s$>$". The input sequence length is constrained to 512. The network output $\tilde{y}_i^n$ of PoE represents the model's confidence about the appropriateness of $r_i^n$ conditioning on $c_i^n$:
\begin{equation*}
\tilde{y}_i^n = p_{\theta_{M}}(y_i^n=1\ |\ c_i^n, r_i^n)
\end{equation*}
where $M$ refers to the PoE model parameterized by $\theta$.






\subsection{Training Objective}
\label{subsec:train-obj}
The training objective of PoE is defined as follows,
\begin{equation*}
\resizebox{\linewidth}{!}{%
$\underset{\theta_M}{\mathrm{argmin}}\E_{\Omega^n\sim\Omega}[\E_{(x_i^n, y_i^n)\in \Omega^n}[-(y_i^n\text{log}\tilde{y}_i^n + (1-y_i^n)\text{log}(1-\tilde{y}_i^n)]]$%
}
\end{equation*}
We minimize the binary cross-entropy loss between $\tilde{y}_i^n$ and $y_i^n$. The whole network is trained in a multitask manner whereby during training, $T$ and $\{\pi^n, \phi^{n}\}_{n=1}^{|N|}$ are jointly optimized. $\{\pi^n, \phi^{n}\}$ are trained to perform their respective domain-specific classification task. More specifically, given a training mini-batch that consists of samples uniformly drawn from any training dataset in $\Omega$, the parameter update of $\{\pi^n, \phi^{n}\}$ only depends on $(x_i^n, y_i^n)\in{\Omega^n}$ in the mini-batch. On the contrary, $T$ is optimized with all training instances in the mini-batch. Hence, $\{\pi^n, \phi^{n}\}$ are domain-specific while $T$ is domain-independent. In this way, $T$ learns to adapt to the multi-domain dialogue dataset and captures a general representation that encodes regularities w.r.t. dialogue data of different domains, while the adapter modules learn to capture the unique characteristics of various dialogue domains, thus serve as the  domain experts.



\subsection{Inference Process}
\label{subsec:inference}
We evaluate a trained PoE on the dialogue evaluation task defined in \S\ref{sec:problem}. Given a context-response input pair $(c_i^j, r_i^j)$ from evaluation dataset $D^j$, PoE will run the forward pass $N$ times in parallel and output $N$ confidence scores denoted as $\{\tilde{y}_i^1,\ldots,\tilde{y}_i^N\}$. The final metric score, $s_i^j$, is computed in the following manner: 

\begin{equation*}
  s_i^j =
    \begin{cases}
      \tilde{y}_i^n & \text{if $D^j$ and $\Omega^n$ share the same domain}\\
      \frac{1}{N}\sum_{t=1}^N{\tilde{y}_i^n} & \text{otherwise}
    \end{cases} 
\end{equation*}
If an evaluation dataset is in-distribution w.r.t. the training data of PoE, we directly apply the confidence score of the corresponding classifier. Here, a single expert makes the decision. For out-of-distribution evaluation datasets, we perform unweighted averaging on all the confidence scores. In this case, all experts jointly make the decision. Hence, our proposed metric is dubbed as PoE: a \textbf{P}anel \textbf{o}f \textbf{E}xperts.  


In out-of-distribution evaluation, the inference involves running the model $N$ times. A simplified strategy is to derive a single adapter and a classifier by averaging the parameters of all $N$ adapters and $N$ classifiers. In practice, we take the arithmetic mean, denoted as  $\phi'$, of the parameters of the individual classifiers $\{\phi^1, \phi^2,\ldots,\phi^N\}$. The parameters of the single adapter module, denoted as $\pi'_l$ for layer $l$, are the arithmetic mean of the parameters of $\{\pi_l^1,\pi_l^2,\ldots, \pi_l^N\}$. $s_i^j$ is obtained with the single module, ($\{\pi'_1,\pi'_2,\ldots,\pi'_{L-1}\}, \phi'$). We justify the use of the arithmetic mean of the parameters of the adapters and classifiers to form a single model by considering the fact that all the adapters share the same configuration as in ~\cite{pmlr-v97-houlsby19a}. Furthermore, the weights of the adapters and classifiers are all initialized with the same uniform distribution. 

We denote PoE after parameter averaging as PoE-avg. PoE-avg combines the domain-specific knowledge of multiple experts (adapters) into a single expert (the adapter after parameter averaging) in an early-fusion manner. It serves as a light-weight variant of PoE and is expected to perform on par with PoE in both in-domain and out-of-domain evaluation.



\subsection{Few-shot Transfer Learning}
\label{subsec:fewshot-transfer}
We also consider a transfer learning setup whereby PoE-avg is finetuned with a small number of human-annotated instances from the target dialogue evaluation dataset. The reason for conducting few-shot transfer learning on PoE-avg instead of PoE is that we often do not have prior knowledge on new dialogue evaluation datasets. Hence, we may need to finetune all adapters in PoE instead of a domain-relevant one. With PoE-avg, we just need to conduct the transfer learning with a single-adapter setup. 

Since the data instances are annotated with continuous human ratings rather than discrete labels, we adopt mean squared error as the optimization objective. Details on the setup of few-shot transfer learning experiments are outlined in \S\ref{subsec:other-consider} 

\section{Multi-domain Dialogue Training Dataset}
\label{sec:dataset}
{The success of the multitask training of PoE relies on a large-scale, high-quality, and multi-domain dialogue dataset. However, it is not trivial to construct such a dataset. In the study of model-based automatic dialogue evaluation, many efforts are devoted to  neural architecture design, however, the development of high-quality training data are not given sufficient attention. There are two major challenges constructing an adequate training dataset.}

{First, existing model-based metrics heavily rely on response sampling strategies~\cite{sinha-etal-2020-learning,zhang-etal-2021-dscore}. Yet, the sampling strategies do not always produce data of good quality. For instance, the commonly-used random utterance selection strategy\footnote{This strategy refers to the random selection of an utterance from a different dialogue as the inappropriate response w.r.t. the current dialogue context.} tends to introduce over-simplistic and false-negative responses. The over-simplistic responses refer to
responses that neither semantically nor syntactically overlap with the dialogue context. As pointed out in ~\cite{zhang2022mddeval}, a large number of such over-simplistic responses mislead the metrics by associating response appropriateness with only content similarity,  thus, introducing unwanted bias. Furthermore, the false negatives can lead the model to misclassify appropriate responses.}

Second, plausible ways to improve the data quality include the implementation human-in-the-loop quality control or creating data with crowd-sourcing. However, these approaches are costly and time-consuming considering that we need a large-scale multi-domain dataset. 

{Overcoming the challenges, by extending~\cite{zhang2022mddeval}, we apply semi-supervised learning techniques to automatically construct a multi-domain context-response dataset through a 3-step workflow, (1) multiple human-human dialogue corpora (\S\ref{subsec:human-corpora}), (2) a set of dialogue response augmentation techniques (\S\ref{subsec:rat}), and (3) a pretrained evaluation model for data pseudo labeling and quality control (\S\ref{subsec:pl}). }

\subsection{Human-Human Dialogue Corpora}
\label{subsec:human-corpora}

Five human-human dialogue corpora are selected for collecting the context-response pairs, as summarized in Table~\ref{tab:data-statistics}, which are  DailyDialog~\cite{li-etal-2017-dailydialog}, ConvAI2~\cite{dinan2020second}, TopicalChat~\cite{gopalakrishnan2019topical}, EmpatheticDialogue~\cite{rashkin-etal-2019-towards} and REDDIT~\cite{NEURIPS2019_fc981212}. We selected the five corpora for the following reasons, (1) They have been used in the studies of open-domain dialogue and are of good quality. (2) Each dialogue corpus is collected with a specific goal, hence in one unique domain.  For instance, ConvAI2 is about persona-guided conversations while DailyDialog focuses on typical topics discussed in our daily life.  (3) They are of a suitable data size for our data collection process.  


\begin{table}[!ht]
\centering
\caption{Human-Human Dialogue Corpora Statistics. We only use the training and validation split of the dialogue corpora, because a part of the test split of the dialogue corpora is used for evaluation in our experiment.}
\resizebox{0.9\linewidth}{!}{
\begin{tabular}{l|cc}
\toprule
\textbf{DailyDialog} & \textbf{training} & \textbf{validation} \\ \midrule
\#dialogues & 11,118 & 1,000 \\
\#utterances & 87,170 & 8,069 \\
\#words & 1,186,046  & 108,933 \\
\#avg utterances per dialogue & 7.84 & 8.07 \\
\#avg words per dialogue & 106.68 & 108.93 \\
\#context-response pairs & 76,052 & 7,069 \\  \midrule
\textbf{EmpatheticDialog} & \textbf{training} & \textbf{validation} \\ \midrule
\#dialogues & 19,529 & 2,768  \\
\#utterances & 84,158 & 12,075 \\
\#words & 1,127,355 & 174,786  \\
\#avg utterances per dialogue & 4.31  & 4.36 \\ 
\#avg words per dialogue & 57.73  & 63.15 \\
\#context-response pairs & 64,629 & 9,307 \\ \midrule
\textbf{ConvAI2} & \textbf{training} & \textbf{validation} \\ \midrule
\#dialogues & 17,878 & 1000 \\
\#utterances & 253,698 & 15,566 \\
\#words & 3,024,032  & 189,374 \\
\#avg utterances per dialogue & 14.19 & 15.57 \\
\#avg words per dialogue & 169.15 & 189.37 \\
\#context-response pairs & 235,820 & 14,566 \\  \midrule
\textbf{TopicalChat} & \textbf{training} & \textbf{validation} \\ \midrule
\#dialogues & 8,627 & 538 \\
\#utterances & 188,357  & 11,660 \\
\#words & 4,374,304 &  273,331 \\
\#avg utterances per dialogue & 21.83 & 21.67 \\ 
\#avg words per dialogue & 507.05 & 508.05 \\
\#context-response pairs & 179,730 & 11,122 \\  \midrule
\textbf{REDDIT} & \textbf{training} & \textbf{validation}  \\  \midrule
\#dialogues & 91,919 & 12,023  \\
\#utterances & 644,429  & 82,927  \\
\#words & 8,104,273 &  1,044,756 \\
\#avg utterances per dialogue & 7.01 & 6.90  \\ 
\#avg words per dialogue & 88.17 &  86.90 \\ 
\#context-response pairs & 523,044 & 65,192 \\ \bottomrule
\end{tabular}
}
\label{tab:data-statistics}
\end{table}

\subsubsection{DailyDialog~\cite{li-etal-2017-dailydialog}} the DailyDialog dataset contains high-quality and human-written conversations that cover a wide range of generic topics, such as relationships, ordinary life, and work. The conversations in DailyDialog are mainly for information exchange and social bond enhancement.

\subsubsection{EmpatheticDialogue~\cite{rashkin-etal-2019-towards}} The EmpatheticDialogue dataset is created for developing dialogue agents that can recognize feelings in the conversation partner and reply accordingly. The conversations are grounded in emotion situations whereby a person describes his or her personal experiences and feelings. The conversation partner acknowledges his or her feelings and then provides appropriate responses.

\subsubsection{ConvAI2~\cite{dinan2020second}} ConvAI2 is an extended dataset of the Persona-Chat~\cite{zhang-etal-2018-personalizing} corpus, which is about exchanging persona information, i.e., the conversations in ConvAI2 are grounded by the personas of the interlocutors. The conversations are about two interlocutors trying to be engaging, to learn about the other's interests, discuss their own interests, and find common ground information~\cite{dinan2020second}. Each persona contains at least 5 sentences describing the corresponding role. In total, there are 1155 possible personas for training. Topic shifts are common within the conversations in ConvAI2 as the interlocutors are continually introducing new information about themselves along the conversation.

\subsubsection{TopicalChat~\cite{gopalakrishnan2019topical}} TopicalChat is a knowledge-grounded human-human conversation dataset, which contains conversations between two interlocutors exchanging knowledge information. The underlying knowledge spans across 8 broad topics, including fashion, politics, books, sports, general entertainment, music, science \& technology, and movies. Each conversation is associated with a Washington Post article and the top three entities by frequency of occurrence in the article. Depending on the configuration, the two interlocutors have access to different knowledge snippets w.r.t. the three entities. The knowledge sources include Wikipedia, Reddit and Washington Post articles.

\subsubsection{REDDIT~\cite{NEURIPS2019_fc981212}} the REDDIT dataset is built on discussions on the Reddit social media platform. There are many different subreddits available, with conversations largely different in topics, language styles, and participation patterns. In total, 109 conversations of at least 3 turns are collected with the median conversation containing 7 utterances. The conversations are extracted from the 2,018 conversational exchanges in the Casual Conversations forum (r/CasualConversations), a community of 607K conversationalists
discussing a variety of topics. Unlike other dialogue corpora, the conversational style in REDDIT is more causal and less organized. In addition, the topics in REDDIT dialogues are more diverse and time-dependent, i.e., the topics may differ a lot when the conversations are carried out at different times.

\subsection{Response Augmentation Techniques}
\label{subsec:rat}

Due to the one-to-many nature of open-ended conversations, the evaluation metrics will benefit from training on multiple appropriate and inappropriate responses per dialogue context. From a human-human dialogue in the five dialogue corpora, we extract 1 to 4 consecutive utterances as the dialogue context, the follow-up utterance serves as the corresponding appropriate response. To generate multiple appropriate and inappropriate responses per context, we need to rely on various response augmentation strategies:

\subsubsection{Syntactic \& Semantic Negative Sampling}
Inspired by~\cite{sinha-etal-2020-learning}, the following perturbation techniques are applied on the original appropriate responses to generate syntactic negative responses: (1) word-drop (a random portion of tokens in the original response, up to 50\%, is dropped). (2) word-shuffle (the order of tokens in the original response is shuffled). (3) word-repeat (randomly repeat words in the original response). For collecting semantic negative responses, we follow the common strategy of randomly sampling an utterance from a different dialogue (within the same corpus) to replace the original appropriate response w.r.t. a dialogue context. To tackle the aforementioned limitations of such random sampling strategy, we leverage additional augmentation techniques as outlined in the subsequent sections and model-in-the-loop quality control measure (\S~\ref{subsec:pl}).  

\subsubsection{Back-Translation}
\label{subsubsec:backtranslation}
Given a dialogue context-response pair extracted from a human-human conversation, Back-Translation~\cite{edunov-etal-2018-understanding} is applied to generate paraphrases of the original response. In the actual implementation, we adopt the pretrained WMT’19 English-German and German-English ensemble model to perform back-translation.

\subsubsection{Generation from State-of-the-art Dialogue Systems}
\label{subsubsec:dialogue-generation}
We rely on state-of-the-art dialogue systems including DialoGPT~\cite{zhang-etal-2020-dialogpt} and BlenderBot~\cite{roller-etal-2021-recipes} to generate a set of appropriate responses with different semantic meanings conditioned on a dialogue context. These systems have been pretrained on a large amount of conversation data and they demonstrate strong ability in generating fluent and on-topic responses.  

\subsubsection{Automatic Generation of Adversarial Responses}
\label{subsubsec:adversarial-sampling}
Motivated by~\cite{gupta-etal-2021-synthesizing}, the mask-and-fill strategy is adopted. There are two steps: (1) masking, where one or a few tokens of a response (up to 15\% of the tokens) are replaced with mask tokens. (2) infilling, a pretrained infilling language model~\cite{donahue-etal-2020-enabling} is adopted to replace the mask tokens in the response with new tokens conditioned on a random dialogue context instead of the original context. For example, if named entities in a response are masked out, the infilling process conditioned on a random context may introduce different named entities that are not consistent with the original context. Such a response may seem to be appropriate in terms of surface lexical features, but in fact, it is semantically inappropriate.

Another strategy is to randomly sample an utterance from the dialogue context and then perform syntactic perturbations on the sampled response. This strategy intends to generate adversarial inappropriate responses that share a certain degree of content similarity with the corresponding contexts. 

Both strategies are intended to automatically construct adversarial negative responses and reduce the reliance on random sampling strategy for introducing semantically negative responses.

\subsection{Pseudo Labeling \& Quality Control}
\label{subsec:pl}
{To avoid excessive false-negative or false-positive data instances in automatically-constructed dataset, we need to put in place a mechanism to filter out low-quality samples. Note that human validation for quality control on a large-scale dataset is costly. We adopt a strong model to provide pseudo labels~\cite{lee2013pseudo} to all the context-response pairs candidates during data augmentation in \S\ref{subsec:rat}. }

Instead of training a model on a human-generated dataset from scratch as what we did in~\cite{zhang2022mddeval}, we leverage the ``Dialogue Evaluator with BERT (DEB)" model released by Sai et al.~\cite{sai-etal-2020-improving}. The rationale is that DEB is first pretrained on a large-scale Reddit dataset (767M Reddit dialogue) and then, finetuned with the human-generated DailyDialog++ dataset. DEB can generalize across domains due to large-scale pretraining while being capable of accurate estimation of response appropriateness due to learning from manually-crafted data. Yeh et al.~\cite{yeh-etal-2021-comprehensive} also proves it to be a strong automatic dialogue evaluator on a large number of turn-level evaluation datasets. 

Concretely, for a context-response pair, DEB provides a soft pseudo label that indicates its confidence on the appropriateness of the response w.r.t. the context. The soft pseudo label is a probability distribution over two classes (appropriate and inappropriate). A confidence threshold of 90\% is adopted to exclude pairs classified by DEB with low confidence.

In the end, for each dialogue corpus, we collected 400K context-response pairs for training and 40K pairs for validation. Both training and validation split are class-balanced, i.e., they contain equal number of appropriate and inappropriate context-response pairs. In total, the multi-domain dataset contains 2M context-response pairs in the training split and 20K pairs in the validation split.

\section{Experiment Preliminaries}
\label{sec:exp-setup}

\begin{table*}[!t]
\centering
    \caption{Summary of the PoE evaluation datasets, that are neither involved in training nor tuning. Part of the information are obtained from~\cite{yeh-etal-2021-comprehensive} and~\cite{chen2021automatic}. * denotes the out-of-domain evaluation datasets w.r.t. the training data. The computation of ``Avg.\#Utts" includes both the context and the response.}
\resizebox{\textwidth}{!}{
    \begin{tabular}{l|cccccccl}
    \toprule
    Name & \#Instances & Avg.\#Utts. & Avg.\#Ctx/Hyp Words & Domain & \#Annotations & Dimension & Neural architecture of the dialogue systems\\
    \midrule
    Persona-USR~\cite{mehri-eskenazi-2020-usr} & 300 & 9.3 & 98.4 / 12.0 & PersonaChat & 5.4K & Maintains Context & Transformer/LSTM Seq2Seq, Memory Network\\
    ConvAI2-GRADE~\cite{huang-etal-2020-grade} & 600 & 3.0 & 24.4 / 11.3 & PersonaChat & 3K & Relevance & Transformer Seq2Seq, DialoGPT, BERT/Transformer Ranker  \\
    Persona-Zhao~\cite{zhao-etal-2020-designing} & 900 & 5.1 & 48.8 / 11.5 & PersonaChat & 3.6K & Appropriateness & Random Sampling, LSTM Seq2SeqAttn, and GPT-2\\
    Persona-DSTC10~\cite{chen2021automatic} & 4,829 & 4.0 & 36.00 / 11.6  & PersonaChat & 77K & Appropriateness & LSTM Seq2SeqAttn, BlenderBot, DialoGPT and GPT-3 \\
    DailyDialog-GRADE~\cite{huang-etal-2020-grade} & 300 & 3.0 & 26.0 / 10.8  & DailyDialog & 3K & Relevance & Transformer Seq2Seq/Ranker \\
    DailyDialog-Zhao~\cite{zhao-etal-2020-designing} & 900 & 4.7 & 47.5 / 11.0 & DailyDialog & 14.4K & Appropriateness & Random Sampling, LSTM Seq2SeqAttn, and GPT-2\\
    DailyDialog-Gupta~\cite{gupta-etal-2019-investigating} & 500 & 4.9 & 49.9 / 10.9 & DailyDialog & 2.5K & Overall & LSTM Seq2Seq, Conditional VAE\\
    Topical-USR~\cite{mehri-eskenazi-2020-usr} & 360 & 11.2 & 236.3 / 22.4 & TopicalChat & 6.5K & Maintains Context & Transformers Seq2Seq\\
    Topical-DSTC10~\cite{chen2021automatic} & 4,500 & 4.0 & 50.6 / 15.9  & TopicalChat & 72K & Appropriateness & LSTM Seq2SeqAttn, BlenderBot, DialoGPT and GPT-3\\
    Empathetic-GRADE~\cite{huang-etal-2020-grade} & 300 & 3.0 & 29.0 / 15.6 & Empathetic & 3K & Relevance & Transformer Seq2Seq/Ranker  \\
    Reddit-DSTC7~\cite{Galley2019GroundedRG} & 9,990 & 3.5 & 35.3 / 11.2 & Reddit & 30K & Relevance & RNN/LSTM Seq2Seq, Memory Network, Pointer-generator\\
    FED-Turn~\cite{mehri-eskenazi-2020-unsupervised}*  & 375 & 10.4 & 87.3 / 13.3 & Other & 17K & Relevance & Meena, Mitsuku\\
    HUMOD~\cite{app10030762}* & 9,500 & 3.9 & 17.0 / 6.1  & Other & 57K & Relevance & Random Sampling\\
    ESL~\cite{lee2020evaluation}* & 1242 & 2.0 & 7.05 / 11.81  & Other & 13K & Overall & BlenderBot, DialoGPT, HRED, Transformer/LSTM Seq2Seq  \\
    NCME~\cite{lee2020evaluation}* & 2461  & 2.0 & 7.34 / 8.57  & Other & 33K & Overall & BlenderBot, DialoGPT, HRED, Transformer/LSTM Seq2Seq \\
    ConTurE~\cite{ghazarian2021user,gunasekara2020overview}* & 1066 & 3.8 & 21.67 / 10.99  & Other & 3.2K & Overall & State-of-the-art systems including Plato and DialoGPT \\
    \bottomrule
    \end{tabular}
}
    \label{tab:eval-data}
\end{table*}

\subsection{Evaluation Datasets}
\label{subsec:eval-task}
We assess PoE on 16 dialogue evaluation datasets. The selection of evaluation datasets is guided by the recent comprehensive survey on automatic dialogue evaluation metrics~\cite{yeh-etal-2021-comprehensive} as well as the "Automatic Evaluation" shared task of DSTC10\footnote{The Tenth Dialog System Technology Challenge (DSTC10)}~\cite{chen2021automatic}. Table~\ref{tab:eval-data} summarizes the essential characteristics of all evaluation datasets. Some evaluation datasets contain annotations along multiple evaluation criteria. For example, the FED-Turn~\cite{mehri-eskenazi-2020-unsupervised} dataset contains annotations along 9 different fine-grained criteria, such as relevance, interestingness, fluency, etc. Since multi-dimensional evaluation is beyond the scope of this work, we only consider response appropriateness in our analysis. For evaluation datasets without annotations along response appropriateness, the criterion that is closest to response appropriateness is considered, such as context relevance or overall quality. As shown in Table~\ref{tab:eval-data}, the ``Dimension" column contains the criteria we consider for our correlation analysis. 

Moreover, the evaluation datasets can be categorized with their respective dialogue domains (as indicated in the ``Domain" column). Those belonging to the ``Other" domain are considered out-of-domain evaluation datasets while the rest are the in-domain evaluation datasets.


\subsection{Baselines}
\label{subsec:baselines}
We compare PoE with three types of systems: (1) published state-of-the-art automatic dialogue evaluation metrics. Specifically, we pick the top-ranked ones that are presented in the comprehensive survey~\cite{yeh-etal-2021-comprehensive}. We include DEB~\cite{sai-etal-2020-improving}, USL-H~\cite{phy-etal-2020-deconstruct}, GRADE~\cite{huang-etal-2020-grade} and USR~\cite{mehri-eskenazi-2020-usr}. USL-H and USR target multi-dimensional evaluation. Hence, both of them contain multiple models with each focus on a specific dialogue quality. Since we only target the response appropriateness, the USR-DR component of USR and the BERT-NUP component of USL-H are adopted respectively. Additionally, we also report the results of the best team in the Track 5.1 of the DSTC10 shared task~\cite{chen2021automatic}. The rationale is that the shared task proposes a meta-evaluation benchmark that covers all evaluation datasets used in this paper except ConTurE, which is more recent than the rest. We want to benchmark PoE against the most recent state-of-the-art automatic dialogue evaluation metric. 

(2) To showcase the advantages brought by the PoE metric alone, instead of the multi-domain training dataset we have collected, we compare PoE against a strong single-model baseline that is trained on the same multi-domain dataset as PoE. The model consists of a pretrained transformer encoder (same as PoE) and a single feed-forward classification network. We denote the baseline as Single-T. In fact, Single-T is similar to the recently proposed MDD-Eval metric~\cite{zhang2022mddeval} since both have the same architecture. Their differences are: (a) Single-T is trained on more data compared with MDD-Eval (2M vs 600K). (b) MDD-Eval is optimized with three different losses while Single-T is optimized with only the cross-entropy loss. We empirically find that Single-T performs on par with MDD-Eval. On some evaluation datasets, it even outperforms MDD-Eval. Hence, we include Single-T instead of MDD-Eval as a baseline in our experiment. 

(3) {To show the advantage of a Panel of Experts (PoE), we compare PoE with a collection of individual domain-specific models, that is referred to as a Collection of Experts (CoE). PoE is optimized with multitask learning with a shared architecture across domains, while CoE is not. An individual model in CoE has the same architecture as Single-T, but trained on one domain-specific subset of the multi-domain dataset. }

\subsection{Experiment Setup}
\label{subsec:other-consider}
Following~\cite{yeh-etal-2021-comprehensive}, the results w.r.t. existing state-of-the-art metrics are computed with the best checkpoints released by the authors. For Single-T, CoE, and PoE, we
repeat the training 10 times with different random seeds to reduce the effect of randomness on model performance. The mean Spearman correlations over the 10 runs are reported for each evaluation dataset. In addition, we perform William's T test~\cite{student1908probable} for pairwise significance tests. In all the tables, we use $\dagger$ on PoE variants if they significantly outperform Single-T, CoE, and MDD-Eval ($p < 0.05$).

We adopt the RoBERTa-base model~\cite{liu2019roberta} as the pretrained transformer encoders in PoE and baselines.This is because RoBERTa has been proven as a powerful text encoder that are beneficial for the automatic dialogue evaluation task in prior works~\cite{zhao-etal-2020-designing,mehri-eskenazi-2020-usr,zhang-etal-2021-dscore,zhang2021investigating}. In addition, we want to have a fair comparison with the existing state-of-the-art metrics, which use either BERT-base or RoBERTa-base except DEB, which is based on BERT-large~\cite{devlin-etal-2019-bert}.

Since the training task for PoE is a binary classification task, we adopt accuracy to determine the model performance. The checkpoint with the best accuracy on average over the five validation datasets is picked to perform the dialogue evaluation task. For the dialogue evaluation task, we adopt Spearman correlations to assess the performance of the ADEs. All experiments are conducted on a single Tesla V100 GPU of 16GB memory. 

Following~\cite{friedman-etal-2021-single}, all our experiments are implemented with PyTorch~\cite{NEURIPS2019_bdbca288}, HuggingFace Transformers~\cite{wolf-etal-2020-transformers} and the adapter-transformers library~\cite{pfeiffer-etal-2020-AdapterHub}. For training PoE, we adopt AdamW optimizer~\cite{loshchilov2018decoupled} with a constant learning rate of 5e-6. We set the training batch size to 32. The number of training epochs is set to 3. The model is evaluated every 2000 steps. If the average validation accuracy does not improve for ten consecutive checkpoints, we stop the training process. In addition, each mini-batch consists of training instances uniformly drawn from each of the training datasets at run time. After the model is fully optimized with multitask training, we freeze the transformer encoder and continue finetuning each adapter separately on their respective training/validation dataset for 10 more epochs. During finetuning, a constant learning rate of 1e-5 is adopted. The model is evaluated every 1024 steps and if the corresponding validation accuracy does not improve for three consecutive checkpoints, we stop the process.

The training hyperparameters of Single-T is the same as those of PoE, except that Single-T doesn't have the adapter finetuning process. For CoE, each domain-specific model is trained exactly in the same manner as Single-T. The only difference is that Single-T is trained on the multi-domain dataset (same as PoE) while the domain-specific models of CoE are trained on their respective in-domain datasets.

We conduct the few-shot transfer learning experiments on both PoE-avg and Single-T. For each evaluation dataset, we randomly sample K\% of the data and K is set to 10\%, 20\%, 30\%, and 40\% respectively. Then, we split the K\% sample set into half with one half for model finetuning and the other half for validation. The target label of each data instance is the average human annotation score. We adopt a training batch size of 2 and set the learning rate to 1e-5. The model is evaluated with Spearman correlation on the validation set. If the correlation doesn't improve for 10 consecutive epochs, we stop the process. After the finetuning process, we evaluate the model on the full evaluation dataset. All the few-shot experiments are repeated 10 times with different random seeds. The analysis in \S\ref{subsec:fewshot-results} is based on the mean Spearman correlations over all the 10 trials.

\begin{table*}[!t]	
\centering
\caption{Spearman correlations (\%) on 11 in-domain evaluation datasets. Scores with p-values $>$ 0.05 are underlined (indicating statistical insignificance). Row 12-16 correspond to the domain-specific average Spearman correlations. The best score for each row is highlighted in bold. Team 5 achieves the first place in the DSTC10 ``Automatic Dialogue Evaluation" shared task. PoE-avg denotes PoE using the parameter averaging inference method. Metrics that are accompanied by an asterisk are trained on domain-specific datasets. \textcolor{black}{Metrics that are accompanied by a plus sign are trained on the same data as PoE.} The rest are domain-independent metrics. $\dagger$ denotes that PoE variants significantly outperforms Single-T, CoE-T, and MDD-Eval ($p < 0.05$).}
\resizebox{\linewidth}{!}{
\begin{tabular}{c|l|ccccccccccc|cc}
\toprule
\textbf{Row} & \textbf{Datasets} & \textbf{USL-H*} & \textcolor{black}{\textbf{USL-H+}} & \textbf{GRADE*} & \textcolor{black}{\textbf{GRADE+}} & \textbf{USR*} & \textcolor{black}{\textbf{USR+}} & \textbf{DEB} & \textbf{Team 5} & \textcolor{black}{\textbf{MDD+}} &\textbf{CoE+} & \textbf{Single-T+} & \textbf{PoE-avg} & \textbf{PoE}  \\ \midrule
1 & Persona-USR & 47.57 & \textcolor{black}{54.02} & 38.31 & \textcolor{black}{42.98} & 56.25 & \textcolor{black}{46.06} & 43.52 & 49.63 & \textcolor{black}{55.03} & 56.11  & 60.72 & 61.85$\dagger$ & \textbf{63.23}$\dagger$  \\
2 & Persona-Zhao & 61.83 & \textcolor{black}{61.56} & 57.45 & \textcolor{black}{63.04} & 51.74 & \textcolor{black}{59.63} & 56.99 & 61.32 & \textcolor{black}{59.74} & 66.74 & 66.97 & \textbf{67.36} & 67.30   \\
3 & Persona-DSTC10 & 39.75 & \textcolor{black}{45.30} & 41.48 & \textcolor{black}{44.40} & 38.65 & \textcolor{black}{41.07} & 38.67 & 43.92 & \textcolor{black}{37.74} & 45.26 & 44.90 & 44.84 & \textbf{45.53}  \\
4 & ConvAI2-GRADE & 56.30 & \textcolor{black}{57.79} & 57.05 & \textcolor{black}{58.28} & 48.51 & \textcolor{black}{51.18} & 50.43 & \textbf{58.43} & \textcolor{black}{44.30} & 57.10 & 56.06 & 57.04 & 57.48  \\
5 & DailyDialog-GRADE & \underline{9.13} & \textcolor{black}{19.00} & 25.40 & \textcolor{black}{20.76} & \underline{7.14} & \textcolor{black}{16.38} & 36.29 & 33.42 & \textcolor{black}{28.08}  & 23.04 & 26.45 & \textbf{36.64}$\dagger$ & 34.16$\dagger$  \\
6 & DailyDialog-Gupta & 61.42 & \textcolor{black}{56.43}  & 59.62 & \textcolor{black}{52.00} & 35.76 & \textcolor{black}{53.44}  & 57.86 & \textbf{63.25} & \textcolor{black}{57.13}  & 55.80 & 61.30 & 61.35 & 61.77 \\ 
7 & DailyDialog-Zhao & 53.11 & \textcolor{black}{45.15} & 53.73 & \textcolor{black}{48.93} & 35.41 & \textcolor{black}{47.83}  & 52.23& 57.54 & \textcolor{black}{56.42} & 51.89 & \textbf{58.26} & 58.24 & 56.13 \\ 
8 & Reddit-DSTC7 & 28.34 & \textcolor{black}{38.31} & 34.12 & \textcolor{black}{38.13} & 37.03 & \textcolor{black}{40.61} & 37.85 & 34.34 & \textcolor{black}{40.12} & 39.89 & 42.19 & \textbf{44.41}$\dagger$ & 43.20$\dagger$  \\ 
9 & Topical-USR & 11.88 & \textcolor{black}{35.13} & 13.84 & \textcolor{black}{34.06} & 36.50 & \textcolor{black}{38.24} & 15.14 & 37.61 & \textcolor{black}{\textbf{55.82}} & 41.45 & 41.26 & 41.42  & 43.72\\ 
10 & Topical-DSTC10 & 20.77 & \textcolor{black}{27.30} & 24.59 & \textcolor{black}{28.71} & 27.41 & \textcolor{black}{28.92} & 29.85  & 27.93 & \textcolor{black}{30.52}  & 33.05 & 31.70 & \textbf{33.30} & 33.20 \\ 
11 & Empathetic-GRADE & 33.40 & \textcolor{black}{42.72} & 34.28 & \textcolor{black}{40.19} & 34.08 & \textcolor{black}{44.27} & 39.56 & 30.57 & \textcolor{black}{37.58} & 43.87 & 44.05 & 46.00 & \textbf{46.36}$\dagger$ \\
\midrule
12 & Average (PersonaChat) & 51.36 & \textcolor{black}{54.67} & 48.57 & \textcolor{black}{52.18} & 48.79 & \textcolor{black}{49.49} & 47.40 & 53.33 & \textcolor{black}{49.20} & 56.30 & 57.16 & 57.77 & \textbf{58.39}  \\
13 & Average (DailyDialog) & 41.22 & \textcolor{black}{40.19} & 46.25 & \textcolor{black}{40.56} & 26.10 & \textcolor{black}{39.22} & 48.79 & 51.40 & \textcolor{black}{47.21} & 43.58 & 48.67 & \textbf{52.08}$\dagger$ & 50.69  \\
14 & Average (TopicalChat) & 16.32 & \textcolor{black}{31.21} & 19.22 & \textcolor{black}{31.39} & 31.95 & \textcolor{black}{33.58} & 22.49 & 32.77 & \textcolor{black}{43.17} & 37.25 & 36.48 & 37.36 & 38.46 \\
15 & Average (Empathetic) & 33.40 & \textcolor{black}{42.72} & 34.28 & \textcolor{black}{40.19} & 34.08  & \textcolor{black}{44.27} & 39.56 & 30.57 & \textcolor{black}{37.58} & 43.87 & 44.05 & 46.00 & \textbf{46.36}$\dagger$ \\
16 & Average (Reddit) & 28.34 & \textcolor{black}{38.31} & 34.12 & \textcolor{black}{38.13} & 37.03 & \textcolor{black}{40.61} & 37.85 & 34.34 & \textcolor{black}{40.12} & 39.89 & 42.19 & \textbf{44.41}$\dagger$ & 43.20$\dagger$ \\
\bottomrule
\end{tabular}}
\label{tab:correlation-in-domain}
\end{table*}

\section{Results \& Analysis}
\label{sec:exp-results}
We would like to answer the following questions: (1) How does PoE perform for in-domain data (\S\ref{subsec:indo-performance})?  (2) How does PoE generalize to out-of-domain evaluation (\S\ref{subsec:ood-performance})? (3) How does PoE  adapt to unseen domains with few-shot transfer learning (\S\ref{subsec:fewshot-results})? 

\subsection{In-Domain Performance of PoE}
\label{subsec:indo-performance}

First, USL-H, GRADE, and USR are domain-specific metrics as they are trained on specific dialogue corpora. USL-H and GRADE are trained on context-response pairs that are based on DailyDialog~\cite{li-etal-2017-dailydialog} while USR are trained on context-response pairs that are derived from TopicalChat~\cite{gopalakrishnan2019topical}. In Table~\ref{tab:correlation-in-domain}, it can be observed that USR performs significantly better on average across all the TopicalChat-related evaluation datasets (row 14) than USL-H and GRADE. On the other hand, USL-H and GRADE perform significantly better than USR on average across all the DailyDialog-related evaluation datasets (row 13). PoE can outperform USL-H and GRADE on the DailyDialog domain as well as USR on the TopicalChat domain. In addition, PoE also outperforms CoE, which are trained on domain-specific datasets, across all the five domains (row 12-16). These observations confirm that PoE is effective for the multi-domain evaluation task. 

\textcolor{black}{Second, to provide a direct and fair comparison between PoE and the domain-specific baselines (which are trained on single-domain data, while PoE is trained on multi-domain data), we re-train USL-H, GRADE, and USR on the same multi-domain data as PoE and present their in-domain performance in columns USL-H+, GRADE+, and USR+ of Table~\ref{tab:correlation-in-domain} respectively. It can be observed that the correlation scores of all three baselines generally improve over their respective domain-specific variants. This shows that better multi-domain evaluation performance can be partially attributed to better training data. However, PoE still outperforms USL-H+, GRADE+, and USR+ by a large margin. The large performance gap is due to (1) the specialized adapter modules of PoE that captures the unique characteristics of different dialogue domains. (2) PoE supports multi-task training, which better exploits the domain-specific features than single-model training. The large performance gap also empirically proves that the dedicated architecture of PoE is essential to the multi-domain dialogue evaluation task.}

Third, we compare PoE with DEB. DEB is a classification model with BERT-large as the backbone. It is first pre-trained on roughly 767M Reddit conversations, then fine-tuned on the DailyDialog++ dataset. Hence, DEB has better generalizability than USL-H, GRADE and USR. PoE outperforms DEB across all five domains. Especially for the Reddit domain (row 16), PoE attains a remarkable improvement of 5.35\% over DEB even though DEB has been pretrained on a large amount of Reddit data. The significant improvement maybe due to that (1) DEB has been applied in our dataset construction process (\S\ref{subsec:pl}). PoE acquires the knowledge of DEB on dialogue evaluation by training on its pseudo-labeled data; (2) Though DEB has more trainable parameters than PoE (340M vs 132M), and is pretrained on a much larger dataset (767M vs 2M),  PoE benefits from its network architecture that captures both  general knowledge across domains, and domain-specific knowledge. 


Additionally, when compared to CoE, PoE attains superior performance across all the domains. The superior performance can be attributed to multitask learning, which serves as an effective tool for boosting model performance through implicit data augmentation and regularization. During the multitask training, the RoBERTa encoder shared by all the adapters in PoE provides general and useful representations for the dialogue contexts and their corresponding responses. Besides the superior performance, PoE is also much more light-weight than CoE in terms of trainable parameters (132M vs 623M).     


\textcolor{black}{Furthermore, PoE also outperforms Single-T and MDD-Eval across all the dialogue domains.} Since both Single-T and PoE are trained on the same data and both share the same type of pretrained transformer encoder (RoBERTa-base), the performance improvement is attributed to PoE's incorporation of the adapters and multitask learning. Even though both Single-T and PoE possess the general knowledge of the multi-domain dialogue data, the different adapter modules in PoE help capture the additional domain-specific knowledge. Moreover, if we compare PoE-avg with Single-T, it can be observed that even though both have approximately the same amount of trainable parameters, PoE-avg achieves better performance than Single-T. The observations confirm that PoE is a more superior multi-domain automatic dialogue evaluation model than Single-T. \textcolor{black}{MDD-Eval performs generally well across the five different domains. Yet, it faces the same limitation as Single-T: lack of dedicated network modules to capture additional domain-specific knowledge. Hence, it performs worse than PoE in the in-domain setting.}




Additionally, PoE variants can outperform the best team (team 5) in the DSTC10 "Automatic Dialogue Evaluation" shared task~\cite{chen2021automatic} on 9 out of 11 evaluation datasets. Remarkably, for the Empathetic (row 15) and the Reddit (row 16) domains, PoE achieves performance gain of approximately 16\% and 9\% respectively in comparison to team 5. Team 5 employs a metric ensemble approach whereby the prediction scores of five different metrics are combined with weighted averaging. The five metrics target relevance, fluency, engagement, specificity, and topic coherence respectively. When performing evaluation on a particular dataset, the weight of each metric is dynamically determined by the Spearman correlation of the metric scores and the corresponding human annotation scores over a subset of that dataset. Different from Team 5's method, PoE doesn't rely on human annotation scores. In addition, it is a single-model metric instead of an ensemble of multiple metrics. 

Moreover, the performance of PoE and PoE-avg is almost identical. This finding confirms our expectation about PoE-avg in \S~\ref{subsec:inference}. It can also be observed that PoE slightly outperforms PoE-avg on the PersonaChat, TopicalChat, and Empathetic domains (row 12, 14, and 15) while PoE-avg slightly performs better than PoE on the DailyDialog and Reddit domains (row 13, 16). A possible reason is that the dialogues in the DailyDialog and Reddit domains are more informal. The language usage is more common. Hence, the knowledge about other dialogue domains can better transfer to the evaluation of such dialogues, but the reverse may not be the same. As a result, PoE-avg, which carries knowledge about different domains, is more capable of evaluating informal dialogues than PoE, which provides domain-specific predictions. 

Lastly, we demonstrate the effectiveness of our data collection pipeline by comparing CoE with USL-H and USR. Concretely, USL-H is the domain expert in evaluating DailyDialog-related dialogues while USR is the domain expert in evaluating TopicalChat-related dialogues. Both USL-H and USR are trained with dialogue context-response data that are developed with the semantic negative sampling strategy introduced in \S\ref{subsec:rat}. There is no quality control on the training data of USR and USL-H. On the other hand, CoE, a collection of domain-specific models that have similar architecture as USL-H and USR, is trained with context-response data that are obtained after our quality control process. We can observe in Table~\ref{tab:correlation-in-domain} that CoE outperforms USL-H by roughly 2\% in terms of the average Spearman correlation over all DailyDialog-related evaluation datasets (row 13). It outperforms USR by roughly 4.5\% in terms of the average Spearman correlation over all TopicalChat-related evaluation datasets (row 14).

\begin{table*}[!t]	
\centering
\small
\caption{Spearman correlations (\%) on 5 out-of-domain evaluation datasets. All the scores are statistically significant. The best score for each row is highlighted in bold. Row 6 corresponds to the average Spearman correlations across all datasets except Team 5 of which the average Spearman correlation is computed with the first four datasets. Metrics that are accompanied by an asterisk are trained on domain-specific datasets. \textcolor{black}{Metrics that are accompanied by a plus sign are trained on the same data as PoE.} The rest are domain-independent metrics. $\dagger$ denotes that PoE variants significantly outperforms variants significantly outperforms Single-T, CoE-T, and MDD-Eval ($p < 0.05$).}
\resizebox{\linewidth}{!}{
\begin{tabular}{c|l|ccccccccccc|cc}
\toprule
\textbf{Row} & \textbf{Datasets} & \textbf{USL-H*} & \textcolor{black}{\textbf{USL-H+}} & \textbf{GRADE*} & \textcolor{black}{\textbf{GRADE+}} & \textbf{USR*} & \textcolor{black}{\textbf{USR+}} & \textbf{DEB} & \textbf{Team 5} & \textcolor{black}{\textbf{MDD+}} &\textbf{CoE+} & \textbf{Single-T+} & \textbf{PoE-avg} & \textbf{PoE}  \\ \midrule
1 & FED-Turn & 19.42 & \textcolor{black}{19.11} & 14.79 & \textcolor{black}{17.97} & 18.32 & \textcolor{black}{31.33} & 18.09 & 29.85 & \textcolor{black}{26.60} & 32.55 & 36.21 & 36.26 &\textbf{36.74}  \\
2 & HUMOD & 64.39 & \textcolor{black}{63.80} & 56.47 & \textcolor{black}{64.11} & 46.08 & \textcolor{black}{59.86} & 64.88 & 64.84 & \textcolor{black}{53.12} & 66.02 & 65.49 & 67.34& \textbf{67.35}$\dagger$  \\
3 & ESL & 34.53 & \textcolor{black}{36.94} & 30.01& \textcolor{black}{33.61} &11.25 & \textcolor{black}{38.59} & 42.24 & 40.01 & \textbf{\textcolor{black}{46.76}}  & 45.95 & 40.24 & 42.94 & 46.37  \\
4 & NCME & 28.02 & \textcolor{black}{28.45} & 21.86 & \textcolor{black}{26.15} & 5.23 & \textcolor{black}{22.24} & 28.57 & 29.60 & \textcolor{black}{21.81} & 19.90 & 26.21 & \textbf{30.55}$\dagger$ & 28.46 \\
5 & ConTurE & 20.80 & \textcolor{black}{25.14} & 24.85& \textcolor{black}{22.31} &23.57& \textcolor{black}{30.82} &30.12 & - & \textcolor{black}{30.39} & 31.16 & 33.16 & 34.74 & \textbf{35.36}$\dagger$ \\
\midrule
6 & Average & 33.43 & \textcolor{black}{34.69} & 29.60& \textcolor{black}{32.38} &20.89& \textcolor{black}{36.57} &36.78 & 41.08 & \textcolor{black}{35.74} & 39.12 & 40.26 & 42.37 & \textbf{42.86}$\dagger$ \\
\bottomrule
\end{tabular}}
\label{tab:correlation-out-of-domain}
\end{table*}

\subsection{Out-of-Domain Evaluation with PoE}
\label{subsec:ood-performance}

A major limitation of existing model-based dialogue evaluation metrics is their inability to generalize to new domains beyond the training data. This is evidenced by the out-of-domain performance of USL-H, USR, and GRADE in Table~\ref{tab:correlation-in-domain}. For example, USL-H and GRADE perform poorly on the TopicalChat domain while USR perform poorly on the DailyDialog domain\footnote{DailyDialog and TopicalChat have the least overlap in characteristics. One focuses on daily conversations while the other targets knowledge exchanges.}. 


PoE has a better out-of-domain generalization ability than the existing metrics as evidenced by its strong Spearman correlations on the five out-of-domain evaluation datasets in Table~\ref{tab:correlation-out-of-domain}. PoE achieves the state-of-the-art performance of 42.86\% (row 6). \textcolor{black}{It can be observed that domain-specific metrics: USL-H*, GRADE*, and USR* perform poorly on out-of-domain evaluation datasets. Even though their generalization to unseen datasets can be improved by training on multi-domain data (as shown in columns: USL-H+, GRADE+, and USR+ in Table~\ref{tab:correlation-out-of-domain}), there is still a significant performance gap between PoE and the domain-specific metrics, which can be attributed to the adoption of multi-task learning in PoE.} 

DEB generally outperforms the domain-specific metrics. This is attributed to its pretraining on large-scale Reddit conversations and the incorporation of a large pre-trained language model (BERT-large). Single-T and CoE is comparable to Team 5, the top team in the DSTC10 ``Automatic Dialogue Evaluation" shared task. The good performance of Single-T and CoE is due to the high-quality multi-domain dataset we have collected. In addition, CoE generalizes to out-of-domain evaluation through ensemble of prediction scores of multiple models while Single-T generalizes through adaptation on the multi-domain dataset.

\textcolor{black}{PoE and PoE-avg significantly outperform the strong baselines, CoE, Single-T, and MDD-Eval} in three out of the five evaluation datasets. Remarkably, PoE can even generalize to the ConTurE dataset~\cite{ghazarian2021user}, which contains dialogue responses generated by ten different state-of-the-art chatbots from the DSTC9 challenge~\cite{gunasekara2020overview}. The evaluation task on ConTurE is difficult as the differences in quality among state-of-the-art chatbots can be quite subtle.

{The strong out-of-domain performance of PoE is attributed to two factors. First, similar to Single-T, PoE is trained on a multi-domain dataset in a multitask manner. It learns to capture the regularities within data across domains. The learned knowledge can effectively transfer to unseen dialogue domains. Second, PoE further exploits the advantage of ensemble models, yet in a light-weight manner. In machine learning, we often adopt ensemble models instead of single model for robust performance. We note that both PoE and CoE  take average of multiple prediction scores for out-of-domain data. However, PoE exploits the light-weight adapters as the hyper-parameters dedicated to effective model generalization, while CoE seeks to generalize with multiple full-fledged domain-specific models.}

Finally, the out-of-domain performance difference of PoE and PoE-avg is insignificant as evidenced by their average Spearman correlations across the five out-of-domain datasets (42.86\%  vs 42.37\%). This reinforces the claim in \S~\ref{subsec:inference} that PoE-avg serves as a light-weight alternative of PoE. 

\begin{table*}[!ht]
\caption{Spearman correlation scores (\%) of Single-T and PoE-avg after few-shot transfer learning on the 16 evaluation datasets when K = 10\%, 20\%, 30\%, and 40\%. The few-shot experiments are performed based on two specific checkpoints of PoE-avg and Single-T respectively, which are trained with the same random seed. The ``Before Finetuning" results correspond to the two specific checkpoints of PoE-avg and Single-T accordingly. Other scores are the mean correlations over 10 different trials. The best score for each dataset is highlighted in bold. All the scores are statistically significant. $\dagger$ denotes that PoE-avg significantly outperforms Single-T ($p < 0.05$) after few-shot transfer learning. Datasets that are accompanied by an asterisk are the out-of-domain evaluation datasets.}
\centering
\resizebox{\linewidth}{!}{
\begin{tabular}{@{}clcccccccccccccc@{}}
\toprule
\multicolumn{2}{c}{} & \multicolumn{2}{c}{Before Finetuning} & & \multicolumn{2}{c}{K = 10\%}  && \multicolumn{2}{c}{K = 20\%} & &
\multicolumn{2}{c}{K = 30\%} & & \multicolumn{2}{c}{K = 40\%}
\\\cmidrule{3-4} \cmidrule{6-7} \cmidrule{9-10} \cmidrule{12-13} \cmidrule{15-16}
Row & Dataset  & Single-T & PoE-avg &  & Single-T & PoE-avg & & Single-T & PoE-avg  & & Single-T & PoE-avg & & Single-T & PoE-avg \\ \midrule
1& Persona-USR & 60.33 & 60.64 && 60.25 & 60.61 && 62.00 & 60.64 && 65.12 & 63.46 && \textbf{67.77} & 64.84\\
2&Persona-Zhao & 68.57 & 68.57 && 68.79 & 68.66 && 71.78 & 69.21 && 74.02 & 73.67 && \textbf{75.92} & 75.35\\
3&Persona-DSTC10 & 44.77 & 44.63 && 47.52 & 50.05$\dagger$ && 49.17 & 54.54$\dagger$ && 51.40 & 57.24$\dagger$  && 54.38 & \textbf{58.48}$\dagger$ \\ 
4&ConvAI2-GRADE & 54.47 & 57.04 && 52.06 & 57.30$\dagger$ && 53.24 & 60.10$\dagger$ && 55.44 & 63.26$\dagger$  &  & 57.15 & \textbf{64.55}$\dagger$ \\ 
5&DailyDialog-GRADE & 26.12 & 36.41 && 39.52 & 41.05 && 48.17 & 45.54 && 53.59 & 53.44 && 58.88 & \textbf{61.21} \\ 
6&DailyDialog-Gupta & 60.06 & 61.23 && 61.93 & 63.57$\dagger$ && 66.52 & 65.42 && 68.52 & 65.88 & & 71.39 & \textbf{73.05}$\dagger$ \\
7&DailyDialog-Zhao & 57.57 & 59.32 && 58.07 & 60.87$\dagger$ && 61.80 & 63.11$\dagger$ && 62.94 & 64.63 &  & 66.50 & \textbf{68.73} \\
8&Reddit-DSTC7 & 42.71 & 42.69 && 46.11 & 49.01$\dagger$ && 49.33 & 51.49$\dagger$ && 49.62 & 52.82$\dagger$ &  & 52.88 & \textbf{53.96} \\ 
9&Topical-USR & 39.89 & 42.36 && 41.99 & 42.14 && 49.29 & 52.92 && 54.11 & 60.99$\dagger$ &  & 60.35 & \textbf{64.51}$\dagger$ \\ 
10&Topical-DSTC10 & 31.18 & 33.14 && 34.40 & 40.32$\dagger$ && 37.68 & 43.87$\dagger$ && 42.23 & 46.92$\dagger$ &  & 42.75 & \textbf{50.34}$\dagger$ \\ 
11&Empathetic-GRADE & 45.33 & 48.01 && 40.23 & 47.30$\dagger$ && 45.56 & 47.71 && 47.85 & 47.28 &  & 48.65 & \textbf{49.05} \\
12&FED-Turn* & 36.07 & 37.29 && 32.28 & 37.86$\dagger$ && 34.11 & 38.07$\dagger$ && 40.42 & 42.15 && 42.75 & \textbf{43.20} \\ 
13&HUMOD* & 65.05 & 66.79 && 68.04 & 68.32 && 69.10 & 69.38 && 70.32 & 71.17$\dagger$ && 71.83 & \textbf{72.25} \\ 
14&ESL* & 35.49 & 42.67 && 45.28 & 48.68$\dagger$ && 45.80 & 50.70$\dagger$ && 48.22 & 52.54$\dagger$ && 51.65 & \textbf{56.42}$\dagger$ \\ 
15&NCME* & 29.37 & 29.96 && 15.17 & 30.13$\dagger$ && 26.13 & 30.04$\dagger$ && 28.42 & 37.63$\dagger$ && 35.92 & \textbf{43.62}$\dagger$ \\
16& ConTurE* & 33.57 & 34.78 && 40.74 & 46.24$\dagger$ && 47.80 & 55.62$\dagger$ && 52.16 & 58.56$\dagger$ && 54.89 & \textbf{62.91}$\dagger$ \\
\midrule
17&Average (PersonaChat) & 57.03 & 57.72 && 57.15 & 59.16$\dagger$ && 59.05 & 61.12$\dagger$ && 61.49 & 64.41$\dagger$ && 63.81 & \textbf{65.81}$\dagger$ \\
18&Average (DailyDialog) & 47.92 & 52.32 & & 53.17 & 55.16$\dagger$ && 58.83 & 58.02 && 61.68 & 61.16 && 65.59 & \textbf{67.67}$\dagger$ \\
19&Average (TopicalChat) & 35.53 & 37.75 & & 38.20 & 41.23$\dagger$ && 43.48 & 48.39$\dagger$ && 48.17 & 53.95$\dagger$ && 51.55 & \textbf{57.42}$\dagger$ \\
20&Average (Empathetic) & 45.33 & 48.01 & & 40.23 & 47.30$\dagger$ && 45.56 & 47.71$\dagger$ && 47.85 & 47.28 && 48.65 & \textbf{49.05} \\
21&Average (Reddit) & 42.71 & 42.69 & & 46.11 & 49.01$\dagger$ && 49.33 & 51.49 && 49.62 & 52.82$\dagger$ && 52.88 & \textbf{53.96} \\
22&Average (Other) & 40.97 & 43.00 & & 40.30 & 46.25$\dagger$ && 44.59 & 48.76$\dagger$ && 47.91 & 52.41$\dagger$ && 51.41 & \textbf{55.68}$\dagger$ \\ 
23&Average (All) & 45.99 & 48.07 & & 46.71 & 50.76$\dagger$ && 50.83 & 53.65$\dagger$ && 53.69 & 56.95$\dagger$ && 56.71 & \textbf{60.16}$\dagger$ \\ \bottomrule
\end{tabular}
}
\label{tab:fewshot-table}
\end{table*}

\begin{table*}[!ht]
{\color{black}
\caption{\textcolor{black}{Spearman correlation scores (\%) of MDD-Eval and PoE-avg after few-shot transfer learning on the 16 evaluation datasets when K = 10\%, 20\%, 30\%, and 40\%. The best score for each dataset is highlighted in bold. All the scores are statistically significant. $\dagger$ denotes that PoE-avg significantly outperforms MDD-Eval ($p < 0.05$) after few-shot transfer learning.}}
\centering
\resizebox{\linewidth}{!}{
\begin{tabular}{@{}clcccccccccccccc@{}}
\toprule
\multicolumn{2}{c}{} & \multicolumn{2}{c}{Before Finetuning} & & \multicolumn{2}{c}{K = 10\%}  && \multicolumn{2}{c}{K = 20\%} & &
\multicolumn{2}{c}{K = 30\%} & & \multicolumn{2}{c}{K = 40\%}
\\\cmidrule{3-4} \cmidrule{6-7} \cmidrule{9-10} \cmidrule{12-13} \cmidrule{15-16}
Row & Dataset  & MDD-Eval & PoE-avg &  & MDD-Eval & PoE-avg & & MDD-Eval & PoE-avg  & & MDD-Eval & PoE-avg & & MDD-Eval & PoE-avg \\ \midrule
1& Persona-USR & 54.48 & 60.64 && 54.92 & 60.61$\dagger$ && 55.56 & 60.64$\dagger$ && 59.74 & 63.46$\dagger$ && 60.70 & \textbf{64.84}$\dagger$ \\

2&Persona-Zhao & 59.97 & 68.57 && 64.18 & 68.66$\dagger$ && 64.80 & 69.21$\dagger$ && 68.44 & 73.67$\dagger$ && 71.15 & \textbf{75.35}$\dagger$ \\

3&Persona-DSTC10 & 37.13 & 44.63 && 43.86 & 50.05$\dagger$ && 44.91 & 54.54$\dagger$ && 50.89  & 57.24$\dagger$  && 48.21 & \textbf{58.48}$\dagger$ \\ 

4&ConvAI2-GRADE & 45.31 & 57.04 && 48.32 & 57.30$\dagger$ && 43.36 & 60.10$\dagger$ && 52.22 & 63.26$\dagger$  && 50.64 & \textbf{64.55}$\dagger$ \\ 

5&DailyDialog-GRADE & 18.09 & 36.41 && 26.98 & 41.05$\dagger$ && 30.74 & 45.54$\dagger$ && 32.22 & 53.44$\dagger$ && 43.19 & \textbf{61.21}$\dagger$ \\ 

6&DailyDialog-Gupta & 57.82 & 61.23 && 57.57 & 63.57$\dagger$ && 60.72 & 65.42$\dagger$ && 59.86 & 65.88$\dagger$ & & 63.67 & \textbf{73.05}$\dagger$ \\

7&DailyDialog-Zhao & 53.40 & 59.32 && 57.13 & 60.87$\dagger$ && 56.23 & 63.11$\dagger$ && 60.88 & 64.63$\dagger$ && 62.68 & \textbf{68.73}$\dagger$ \\

8&Reddit-DSTC7 & 28.91 & 42.69 && 36.65 & 49.01$\dagger$ && 37.64 & 51.49$\dagger$ && 40.81 & 52.82$\dagger$ && 44.08 & \textbf{53.96}$\dagger$ \\ 

9&Topical-USR & 51.43 & 42.36 && 61.57 & 42.14 && 59.68 & 52.92 && 58.51 & 60.99$\dagger$ && 64.41 & \textbf{64.51} \\ 

10&Topical-DSTC10 & 28.44 & 33.14 && 28.47 & 40.32$\dagger$ && 38.24 & 43.87$\dagger$ && 38.91 & 46.92$\dagger$ & & 42.97 & \textbf{50.34}$\dagger$ \\ 

11&Empathetic-GRADE & 37.66 & 48.01 && 44.37 & 47.30$\dagger$ && 41.55 & 47.71$\dagger$ && 26.44 & 47.28$\dagger$ && 30.11 & \textbf{49.05}$\dagger$ \\

12&FED-Turn* & 26.52 & 37.29 && 29.08 & 37.86$\dagger$ && 16.73 & 38.07$\dagger$ && 27.49 & 42.15$\dagger$ && 21.97 & \textbf{43.20}$\dagger$ \\ 

13&HUMOD* & 52.08 & 66.79 && 57.53 & 68.32$\dagger$ && 60.15 & 69.38$\dagger$ && 61.10 & 71.17$\dagger$ && 64.23 & \textbf{72.25}$\dagger$ \\ 

14&ESL* & 45.32 & 42.67 && 24.35 & 48.68$\dagger$ && 38.35 & 50.70$\dagger$ && 27.55 & 52.54$\dagger$ && 33.38 & \textbf{56.42}$\dagger$ \\ 

15&NCME* & 22.71 & 29.96 && 22.04 & 30.13$\dagger$ && 10.10 & 30.04$\dagger$ && 12.95 & 37.63$\dagger$ && 24.13 & \textbf{43.62}$\dagger$ \\

16& ConTurE* & 27.76 & 34.78 && 36.76 & 46.24$\dagger$ && 49.63 & 55.62$\dagger$ && 53.56 & 58.56$\dagger$ && 61.89 & \textbf{62.91} \\
\midrule

17&Average (PersonaChat) & 49.22 & 57.72 && 52.82 & 59.16$\dagger$ && 52.16 & 61.12$\dagger$ && 57.82 & 64.41$\dagger$ && 57.68 & \textbf{65.81}$\dagger$ \\

18&Average (DailyDialog) & 43.10 & 52.32 & & 47.23 & 55.16$\dagger$ && 49.23 & 58.02$\dagger$ && 50.99 & 61.16$\dagger$ && 56.49 & \textbf{67.67}$\dagger$ \\

19&Average (TopicalChat) & 39.94 & 37.75 & & 45.02 & 41.23 && 48.96 & 48.39 && 48.71 & 53.95$\dagger$ && 53.69 & \textbf{57.42}$\dagger$ \\

20&Average (Empathetic) & 37.66 & 48.01 & & 44.37 & 47.30$\dagger$ && 41.55 & 47.71$\dagger$ && 26.44 & 47.28$\dagger$ && 30.11 & \textbf{49.05}$\dagger$ \\

21&Average (Reddit) & 28.91 & 42.69 & & 36.65 & 49.01$\dagger$ && 37.64 & 51.49$\dagger$ && 40.81 & 52.82$\dagger$ && 44.07 & \textbf{53.96}$\dagger$ \\

22&Average (Other) & 34.88 & 43.00 & & 33.95 & 46.25$\dagger$ && 34.99 & 48.76$\dagger$ &&  36.53 & 52.41$\dagger$ && 41.12 & \textbf{55.68}$\dagger$ \\ 

23&Average (All) & 40.44 & 48.07 & & 43.36 & 50.76$\dagger$ && 44.27 & 53.65$\dagger$ && 45.72 & 56.95$\dagger$ && 49.21 & \textbf{60.16}$\dagger$ \\ \bottomrule
\end{tabular}
}
\label{tab:fewshot-mdd-table}}
\end{table*}

\subsection{Few-shot Transfer Learning}
\label{subsec:fewshot-results}

Besides its strong in-domain performance and zero-shot generalization, PoE has an additional benefit, the exploitation of few-shot transfer learning for fast adaptation to new dialogue domains. \textcolor{black}{In this section, we analyze the few-shot transfer performance of PoE-avg, Single-T, and MDD-Eval. All the models carry knowledge of dialogues across multiple domains. Yet, the architecture of PoE-avg is more sophisticated due to the incorporation of the adapter. Table~\ref{tab:fewshot-table} presents the performance comparison between Single-T and PoE-avg and Table~\ref{tab:fewshot-mdd-table} shows the comparison between MDD-Eval and PoE-avg.} 

\textcolor{black}{As shown in the two tables, the performance Single-T, MDD-Eval, and PoE-avg generally improves as K increases. An absolute improvement of 12\% is achieved by PoE-avg in terms of the average Spearman correlations over the 16 evaluation datasets when K = 40\% (from 48.07\% to 60.16\%). Single-T and MDD-Eval also attain 10.72\% and 8.77\% absolute improvement respectively when K = 40\%.}


With only 5\% of the data for finetuning (K = 10\%), few-shot transfer learning brings PoE-avg more than 5\% absolute improvement on Persona-DSTC10, Reddit-DSTC7, Topical-DSTC10, and ESL. Remarkably, the improvement on ConTurE is the most significant among all datasets (more than 11\% improvement). \textcolor{black}{Additionally, PoE-avg outperforms Single-T and MDD-Eval by 6\% and 12.3\% respectively in terms of average Spearman correlation across all out-of-domain evaluation datasets (row 22).} The observations reinforces that PoE-avg is capable of fast adaptation to new dialogue domains. 

\textcolor{black}{In general, PoE-avg significantly outperforms both Single-T and MDD-Eval in most of the datasets for all choices of K. The only exceptions are that (1) when K $\geq$ 20\%, Single-T performs better than PoE-avg on Persona-USR and Persona-Zhao; (2) When K $\leq$ 20\%, MDD-Eval outperforms PoE-avg on Topical-USR.} In most cases, the performance of Single-T becomes comparable with that of PoE-avg only when finetuned on more data. This showcases that PoE-avg is better at few-shot transfer and more data-efficient than Single-T.   

With 20\% of the data for finetuning (K = 40\%), PoE-avg can obtain more than 50\% Spearman correlations on 13 out of 16 datasets and more than 60\% Spearman correlations on 9 datasets. Especially on most of the out-of-domain evaluation datasets (rows 12-16), the performance improvement is significant. On average, there is an improvement of 12.68\% Spearman correlations (from 43.00\% to 55.68\%). From the observations, we can conclude that few-shot transfer learning with PoE-avg offers us a scalable way for the multi-domain dialogue evaluation task. This is because whenever we need to evaluate new dialogues, we may just need to annotate a few in-domain data instances and then finetune the pretrained PoE-avg model with the annotated data. Subsequently, we can apply the finetuned model for the new evaluation task.

\subsection{\textcolor{black}{Application to Downstream Dialogue Tasks}}
\label{subsec:apply-drs}

\textcolor{black}{The purpose of developing automatic dialogue evaluation metrics is to benefit downstream dialogue tasks, such as dialogue generation and response selection. With metrics that strongly correlate with human evaluation, researchers and practitioners can accurately estimate the performance of their models during the development phase. Besides providing accurate quality estimation of model responses, PoE can be directly used for response re-ranking or response selection. In this section, we examine whether PoE is an effective response selector on the Persona-Chat benchmark~\cite{zhang-etal-2018-personalizing}.}

\textcolor{black}{The Persona-Chat benchmark consists of 8939 dialogues for training, 1000 for validation, and 968 for testing. Response selection is conducted at every turn of the dialogue. Hence, there are 65719 context-response pairs for training, 7801 for validation, and 7512 for test. For each dialogue context, there is a true positive response, which is the original human response, and 19 distractors. The aim of PoE is to rank the true positive response at the top-1 among the 20 response candidates in terms of contextual relevance. Its effectiveness is measured by the recall of the true positive responses, denoted as hits@1, which is a common metric for evaluating response selection systems. Each dialogue in Persona-Chat is accompanied by two persona profiles and each persona profile consists of 3 to 5 sentences describing the background information of the corresponding interlocutor. There are 955 possible personas for training, 100 for validation, and 100 for test. Since the original persona sentences are similar to the utterances within the dialogues, the authors also provide a set of revised persona profiles. In our experiments, we consider three different settings: (1) response selection without persona profile information (denoted as dialogue-only); (2) response selection with the original persona profile (denoted as original); (3) response selection with the revised persona profile (denoted as revised).}

\textcolor{black}{In addition, we apply PoE for response selection in two different ways. First, directly run inference on the test set with the pre-trained PoE model (denoted as PoE-direct). Second, freeze the transformer encoder of PoE and train a new adapter on the Persona-Chat training data. Then, we can run inference on the test set with the new adapter module (denoted as PoE-adapter). We compare the two PoE variants against different ranking models, which include as baselines the current SotA models for retrieval approaches from~\cite{zhang-etal-2018-personalizing}, the Interactive Matching Network (IMN)~\cite{10.1145/3357384.3358140}, and the Dually Interactive Matching Network (DIM)~\cite{gu-etal-2019-dually}.}

\textcolor{black}{Table~\ref{tab:response-selection} presents the Hits@1 (\%) of PoE variants and previous methods on Persona-Chat. We can make the following observations: (1) even though PoE-direct is not finetuned on the Persona-Chat response selection data, it still performs much better than the ranking baselines under the dialogue-only, original and revised settings. (2) PoE-adapter outperforms IMN in the original and revised settings. Even though PoE-adapter performs slightly worse than DIM, it is much more parameter-efficient. The total number of trainable parameters of DIM is approximately 6.23M while that of PoE-adapter is 1.79M. IMN and DIM require full model training to reach satisfactory performance whereas for PoE-adapter, we only need to train the light-weight adapter module and reuse its prior knowledge obtained from pre-training on the multi-domain dialogue data. In addition, we did not conduct hyperparameter search to optimize PoE-adapter's performance on the downstream response selection task. Instead, we simply reuse the hyperparameters that are applied to the dialogue evaluation task. (3) PoE-adapter significantly outperforms PoE-direct under the three persona configurations. This observation is expected as training on task-specific data always improves the model performance on the corresponding downstream task.} 

\textcolor{black}{The experiment results not only showcase the usefulness of PoE to the response selection task, it also demonstrate the claim that PoE is much more flexible than single-model metrics. When adapting to a new task-specific dataset, we can just incorporate a task-specific light-weight adapter into PoE. The number of trainable parameters is much less compared to full-model finetuning.}




\begin{table}[!t]	
\centering
{\color{black}
\caption{\textcolor{black}{Hits@1 (\%) of PoE variants and previous methods on Persona-Chat under the three persona
configurations.}}
\resizebox{0.9\linewidth}{!}{
\begin{tabular}{l|ccc}
\toprule
\textbf{Models} & \textbf{Dialogue-only} & \textbf{Original} & \textbf{Revised}  \\ \midrule
IR Baseline~\cite{zhang-etal-2018-personalizing} & 21.4 & 41.0 & 20.7 \\
Starspace~\cite{zhang-etal-2018-personalizing} & 31.8 & 48.1 & 32.2  \\
Profile~\cite{zhang-etal-2018-personalizing} & 31.8 & 47.3 & 35.4 \\
KV Profile~\cite{zhang-etal-2018-personalizing} & 34.9 & 51.1 & 35.1  \\ \midrule
IMN~\cite{10.1145/3357384.3358140} & 63.8 & 66.7 & 64.0 \\
DIM~\cite{gu-etal-2019-dually} & 63.8 & 78.8 & 70.7 \\ \midrule
PoE-direct & 51.0 & 57.6 & 49.3 \\
PoE-adapter & 63.6 & 78.0 & 68.3 \\
\bottomrule
\end{tabular}}
\label{tab:response-selection}}
\end{table}

\subsection{\textcolor{black}{Efficiency Analysis}}
\label{subsec:efficiency-analysis}

\textcolor{black}{In this section, we compare the efficiency of PoE and the baselines. In terms of training, given 2 million data instances and a training batch size of 32 (62500 steps per epoch), PoE takes roughly 7.5 hours to complete one epoch on a single Nvidia RTX 3090 GPU card. Both Single-T and CoE take around 5.5 hours per epoch. Since model training only requires three epochs and the early stopping strategy is implemented, the additional training time is worthwhile considering the superior performance of PoE.}

\textcolor{black}{In terms of inference, the speed of Single-T and that of PoE-avg are similar. The difference between PoE and CoE is negligible. However, the inference time of PoE or CoE is N times as that of Single-T or PoE-avg where N denotes the number of domain-specific adapters or domain-specific models. In our experiments, PoE and CoE takes 2.5 minutes to complete inference on 10000 data instances with an evaluation batch size of 32 and a single Nvidia RTX 3090 GPU card. Single-T and PoE-avg take around 30 seconds. Hence, in real-life evaluation task, PoE-avg is the best choice considering the fact that it is not only efficient, but also achieves comparable performance with PoE in terms of correlations with human evaluation.}

\subsection{\textcolor{black}{Beyond PoE-avg}}
\label{subsec:efficiency-analysis}

\textcolor{black}{Besides parameter averaging, we explore other ways to collapse the domain-specific adapters and classifiers into a single adapter and a classifier respectively. More specifically, we examine two other methods, max pooling of the parameters and min pooling of the parameters, which are denoted as PoE-max and PoE-min respectively. Table~\ref{tab:other-than-poeavg} presents the Spearman correlations of PoE-max, PoE-avg, and PoE-min on the 16 dialogue evaluation benchmarks. It can be observed that the three methods are not significantly different in terms of both in-domain (row 17-21) and out-of-domain (row 22) evaluation performance. PoE-min performs slightly worse than PoE-avg or PoE-max. Hence, both PoE-avg and PoE-max can be used as efficient alternatives to PoE during inference.}

\begin{table}[!t]	
\centering
{\color{black}
\caption{\textcolor{black}{Performance comparison of PoE-max, PoE-avg, and PoE-min. Datasets that are accompanied by an asterisk are the out-of-domain evaluation datasets. All the scores are statistically significant.}}
\resizebox{\linewidth}{!}{
\begin{tabular}{c|l|ccc}
\toprule
\textbf{Row} & \textbf{Datasets} & \textbf{PoE-max} & \textbf{PoE-avg} & \textbf{PoE-min}  \\ \midrule
1 &  Persona-USR & 62.15 & 61.85 & 61.91 \\
2 & Persona-Zhao & 66.70 & 67.36 & 65.73  \\
3 & Persona-DSTC10 & 44.98 & 44.84 & 44.45 \\
4 & ConvAI2-GRADE & 56.67 & 57.04 & 55.98  \\
5 & DailyDialog-GRADE & 34.85 & 36.64 & 35.34 \\
6 & DailyDialog-Gupta & 60.69 & 61.35 & 60.54 \\ 
7 & DailyDialog-Zhao & 57.56 & 58.24 & 57.42 \\ 
8 & Reddit-DSTC7 & 43.85 & 44.41 & 43.80  \\ 
9 & Topical-USR & 41.32 & 41.42 & 41.13 \\ 
10 & Topical-DSTC10 & 33.40 & 33.30 & 33.20 \\ 
11 & Empathetic-GRADE & 45.77 & 46.00 & 45.69 \\ \midrule
12 & FED-Turn* & 35.24 & 36.26 & 32.54 \\
13 & HUMOD* & 67.24 & 67.34 & 67.19 \\
14 & ESL* & 43.65 & 42.94 & 41.94 \\
15 & NCME* & 30.38 & 30.55 & 29.99 \\
16 & ConTurE* & 34.59 & 34.74 & 34.04 \\ \midrule
17 & Average (PersonaChat) & 57.63 & 57.77 & 57.02 \\
18 & Average (DailyDialog) & 51.03 & 52.08 & 51.10 \\
19 & Average (TopicalChat) & 37.36 & 37.36 & 37.16 \\
20 & Average (Empathetic) & 45.77 & 46.00 & 45.69 \\
21 & Average (Reddit) & 43.85 & 44.41 & 43.80 \\
22 & Average (Other) & 42.22 & 42.37 & 41.14 \\
23 & Average (All) & 47.44 & 47.77 & 46.93 \\
\bottomrule
\end{tabular}}
\label{tab:other-than-poeavg}}
\end{table}

\section{Conclusions and Future Work}

This paper studies multi-domain dialogue evaluation. We address the poor generalization issue of existing model-based automatic dialogue evaluation metrics, and propose a novel metric, that is called a panel of experts (PoE), with transformer adapters under multitask learning. To facilitate the training of PoE, we further construct a high-quality multi-domain dataset via data augmentation and pseudo labeling. Through extensive and comprehensive experiments on a large collection of evaluation datasets, we demonstrate that PoE strongly correlates with human judgements and outperforms the baselines and existing state-of-the-art evaluation metrics. In addition, PoE exhibits strong zero-shot generalization and few-shot transfer performance. In the future, we will extend PoE for multi-dimensional evaluation by incorporating novel pretraining objectives, such as interestingness and informativeness. In addition, we will extend our study  from turn level evaluation towards dialogue level evaluation. 

\bibliographystyle{IEEEtran}

\end{document}